\documentclass{article}

\usepackage[main, final]{neurips_2026}
\usepackage{graphicx}
\usepackage{subcaption}
\usepackage{enumitem}
\usepackage{wrapfig}

\usepackage[utf8]{inputenc} 
\usepackage[T1]{fontenc}    
\usepackage{hyperref}       
\usepackage{url}            
\usepackage{booktabs}       
\usepackage{amsfonts}       
\usepackage{nicefrac}       
\usepackage{microtype}      

\usepackage{amsmath}
\usepackage{amssymb}
\usepackage{mathtools}
\usepackage{amsthm}
\usepackage{algorithm}
\usepackage{algorithmic}
\usepackage{multirow}
\usepackage{makecell}
\usepackage[capitalize,noabbrev]{cleveref}
\usepackage[table]{xcolor}
\usepackage[most]{tcolorbox}

\usepackage[textsize=tiny]{todonotes}

\definecolor{myBlue}{HTML}{1358be}
\definecolor{myRed}{HTML}{c00000}

\newif\ifcomments
\commentstrue

\title{Progress-Aware Distillation for Mitigating Stagnation\\in Small Language Model Agents}

\author{%
  Yuanpu Cao$^{1}$, Saket Sathe$^{2}$, Hanyu Wang$^{1}$, Ziyi Yin$^{1}$, Fenglong Ma$^{1}$, Jinghui Chen$^{1}$\\
  $^{1}$The Pennsylvania State University
  \quad
  $^{2}$Microsoft CoreAI
}

\begin{document}

\maketitle
\makeatletter
\begingroup
\renewcommand{\thefootnote}{}
\renewcommand{\@makefntext}[1]{\noindent#1}
\footnotetext{Correspondence to: Yuanpu Cao
\texttt{\textless yuanpu.cao@psu.edu\textgreater}
and Jinghui Chen
\texttt{\textless jzc5917@psu.edu\textgreater}.\newline
Code is available at \url{https://github.com/CaoYuanpu/ProD}.}
\endgroup
\makeatother

\begin{abstract}
While LLM-based agent systems have demonstrated remarkable proficiency in reasoning and solving complex real-world tasks, they also come with substantial token consumption and inference latency, limiting their practical deployment. To enhance the efficiency of agent systems, recent efforts have focused on transferring agentic capabilities from large-scale models to small language models (SLMs) via supervised fine-tuning-based distillation. Nevertheless, as interaction trajectories grow longer, involving extended reasoning-action chains and extensive environmental feedback, distilled student agents tend to experience progress stagnation: becoming trapped in unproductive loops characterized by persistent failed strategies and difficulty making meaningful progress. Our systematic analysis shows that this phenomenon consistently occurs across diverse SLMs and task domains. To address this issue, we propose Progress-Aware Distillation (ProD), which explicitly detects and penalizes stagnation in student-generated trajectories. ProD encodes the stagnation gap between student and teacher into an adaptive margin for dynamically penalizing possible stagnation. By iteratively applying this procedure, ProD enables student agents to progressively approximate teacher behavior. Experiments with 8 SLMs, ranging from 0.6B to 8B parameters, on in-domain and out-of-domain benchmarks demonstrate that ProD substantially mitigates progress stagnation while improving success rates and task-completion efficiency.
\end{abstract}
\section{Introduction}
Recent advances in large language models (LLMs) have led to strong generalization across diverse natural language processing tasks \citep{gpt4, qwen3}. To extend beyond conventional language modeling, LLM-powered agents \citep{agent-survey} integrate action modules with controlled API access, enabling the invocation of task-specific tools and interaction with external environments \citep{toolformer}. This agentic design allows LLMs to address more demanding real-world challenges, including web-based information retrieval \citep{webexplorer}, software engineering \citep{sweagent}, and scientific report generation \citep{webthinker, deepresearch}. In these complex settings, LLM agents typically employ iterative Reason--Act--Observe cycles to incrementally make progress and refine solutions \citep{react, codeact}. While agents built on large proprietary backbones such as GPT-4.1 \citep{gpt4.1} achieve high task success rates, their prohibitive token consumption and inference latency pose significant barriers to practical deployment.

To bridge this efficiency gap, transferring agentic capabilities from large-scale models to small language models (SLMs) has emerged as a compelling research direction. Inspired by the success of chain-of-thought (CoT) distillation \citep{fullcotdistill-0, fullcotdistill-1}, a recent agent distillation framework \citep{agentdistill} leverages expert trajectories composed of (Thought, Action, Observation) tuples generated by high-capacity teachers during problem solving. By performing behavioral cloning via supervised fine-tuning (SFT) on these trajectories, SLMs are trained to emulate teacher behaviors and thereby acquire tool-augmented reasoning capabilities.

However, despite familiarizing SLMs with agent execution workflows, this SFT-based distillation strategy suffers from a critical limitation during task execution. Specifically, as agents perform successive Reason--Act--Observe iterations, the runtime context accumulates long reasoning-action chains alongside large volumes of environment-derived feedback, such as noisy web search results and code execution logs. Under these increasingly complex contexts, distilled SLM agents frequently exhibit a pronounced form of \emph{progress stagnation}: they become trapped in unproductive loops, persistently attempting failed strategies and actions from prior turns, and struggle to break these cycles to make forward progress. Our systematic analysis (in \cref{sec:stagnation_analysis}) reveals that this stagnation is pervasive across diverse model families (e.g., Qwen3 \citep{qwen3}, Llama3 \citep{llama3}, and Gemma3 \citep{gemma3}), various task domains, and different fine-tuning paradigms. Moreover, the issue becomes increasingly severe as model scale decreases and trajectory length increases, ultimately creating a fundamental bottleneck for effective agent distillation.

To address this limitation, we propose \textbf{Pro}gress-Aware \textbf{D}istillation (\textbf{ProD}), a novel framework that explicitly detects and penalizes stagnation in student-generated trajectories during the distillation process. To achieve precise intervention, ProD encodes the stagnation gap (i.e., the difference in stagnation levels) between student and teacher trajectories into an adaptive margin. This margin dynamically modulates the optimization strength, enabling aggressive correction of stagnant behaviors while preventing overcorrection on student trajectories that already closely align with teacher progress. By iteratively applying this optimization procedure, ProD substantially alleviates progress stagnation in SLM-based agents, enabling them to steadily converge toward the coherent problem-solving behaviors of high-capacity teacher agents. 

We summarize our contributions as follows.
\begin{itemize}[leftmargin=1.8em]
    \item We identify and systematically analyze progress stagnation, a pervasive failure mode in SFT-based agent distillation where SLMs collapse into unproductive loops during multi-turn interactions. Our analysis reveals that this phenomenon consistently occurs across various model families and task domains.

    \item We propose the Progress-Aware Distillation (ProD) framework, which explicitly detects and penalizes stagnation exhibited in student-generated trajectories during optimization. By encoding student--teacher stagnation gap into an adaptive optimization margin, ProD dynamically penalizes possible stagnation during the distillation process.
    
    \item Comprehensive experiments on 8 SLMs ranging from 0.6B to 8B parameters across  in-domain and out-of-domain benchmarks demonstrate that ProD substantially mitigates progress stagnation while enhancing both task success rates and completion efficiency for distilled SLM agents.
\end{itemize}

\begin{figure*}[t]
\begin{center}
\centerline{\includegraphics[width=1.0\linewidth]{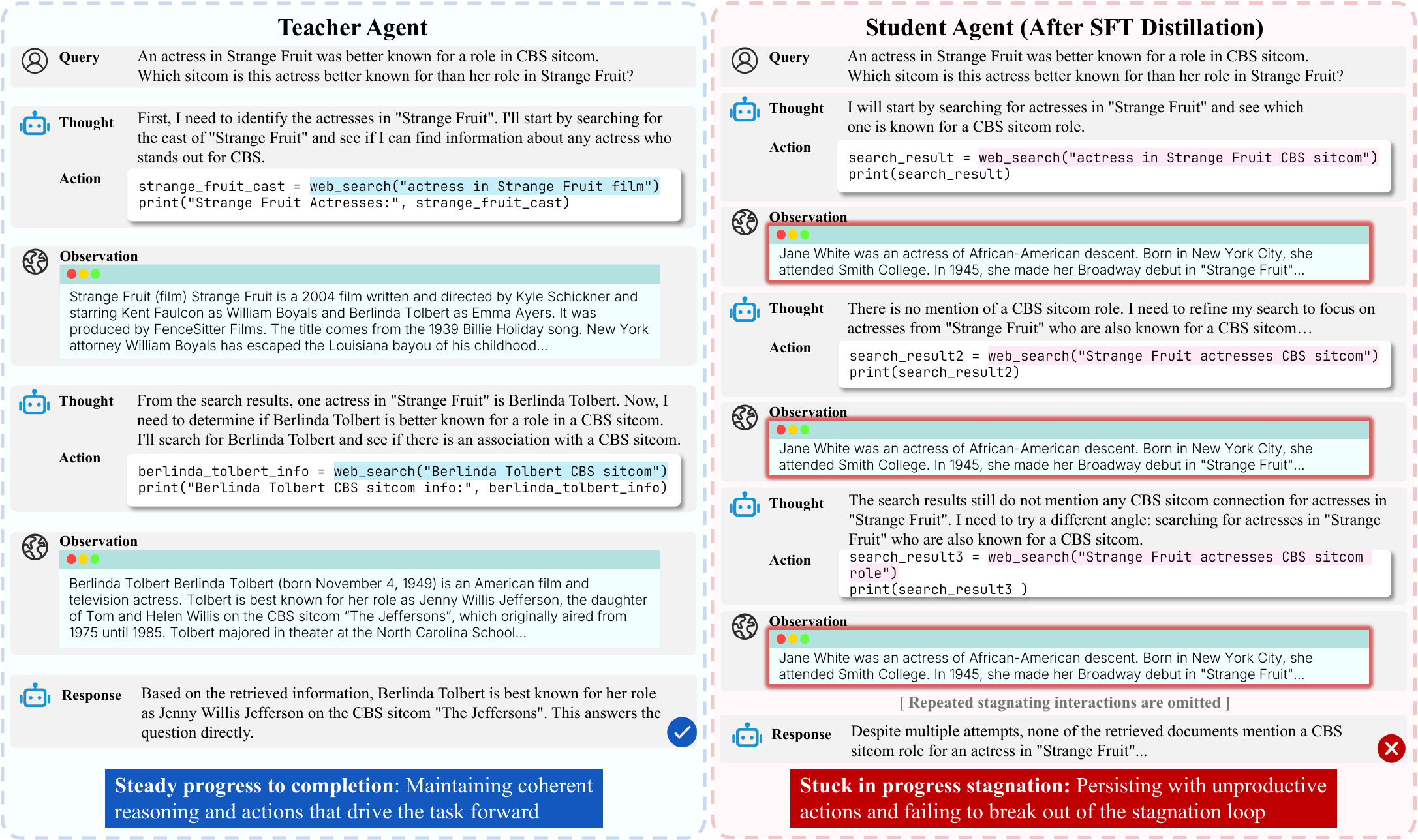}}
\caption{Comparison between the teacher agent (GPT-4.1) and the student agent (SFT-distilled Qwen3-1.7B) on a web information-retrieval task. The teacher maintains steady progress through coherent reasoning and actions \textbf{\color{myBlue}{(left)}}, while the student suffers from progress stagnation, cycling through similar but ineffective search queries and remaining stuck \textbf{\color{myRed}{(right)}}.}
\label{fig:case}
\end{center}
\vskip -0.4in
\end{figure*}
\section{Progress Stagnation Analysis} \label{sec:stagnation_analysis}
We organize this section as follows. In \cref{sec:preliminary}, we review the standard SFT-based agent distillation framework. In \cref{sec:progress_stagnation}, we analyze the progress stagnation phenomenon in SFT-distilled SLM agents and introduce \emph{stagnation scores} to quantify the associated failure modes.
\subsection{Preliminaries: Agent Distillation via SFT} \label{sec:preliminary}
The goal of agent distillation is to transfer agentic capabilities from a powerful teacher model $p_T$ to a compact student model $p_S^{\boldsymbol{\theta}}$. Under multi-turn interaction frameworks such as ReAct \citep{react} and CodeAct \citep{codeact}, agent distillation typically begins by collecting execution trajectories from the teacher model. Each trajectory $\tau $ is a sequence of interaction tuples consisting of thoughts ($\boldsymbol{h}$), actions ($\boldsymbol{a}$), and environmental observations ($\boldsymbol{o}$). Specifically, given a user query $\boldsymbol{x}$ and an agent instruction $\boldsymbol{I}_{\text{agent}}$ (e.g., ``To solve the task, you must plan forward to proceed in a series of steps, in a cycle of Thought, Code, and Observation sequences.'' \citep{smolagents}), the teacher produces a trajectory 
\(
\tau = \big[\, 
(\boldsymbol{h}_{1}, \boldsymbol{a}_{1}, \boldsymbol{o}_{1}),
\ldots,
(\boldsymbol{h}_{L_\tau}, \boldsymbol{a}_{L_\tau}, \boldsymbol{o}_{L_\tau})
\,\big]
      \sim p_{T}(\cdot \mid \boldsymbol{x}, \boldsymbol{I}_{\text{agent}})
\), where $L_\tau$ denotes the total number of interaction turns. At each turn $i$, the agent generates a reasoning thought $\boldsymbol{h}_i$ and an action $\boldsymbol{a}_i$ to invoke external tools. For instance, CodeAct employs executable code as the action space, providing greater flexibility and broader software compatibility than the text- or JSON-based formats used in the conventional ReAct framework. The observation $\boldsymbol{o}_i$ represents the environment's response to the action $\boldsymbol{a}_i$ and is not generated by the model. Following the data collection procedures in prior research \citep{fireact, tora}, \citet{agentdistill} constructs the distillation dataset $\mathcal{D}$ by filtering and retaining only successful  teacher trajectories that lead to the correct final answer. The student agent is then trained via SFT, with the loss calculated exclusively on teacher-generated tokens while masking out observations. For notational convenience, we define the log-likelihood of a trajectory (excluding observations) as \(
\log p_S^{\boldsymbol{\theta}}(\tau \mid \boldsymbol{x})
:=
\sum_{i=1}^{L_\tau}
\log p_S^{\boldsymbol{\theta}}(\boldsymbol{h}_{i}, \boldsymbol{a}_{i} \mid \boldsymbol{x}, \tau_{<i})
\), where $\tau_{<i}$ denotes the interaction history prior to turn $i$. The SFT objective is thus \(  \mathcal{L}_{\text{SFT}}({\boldsymbol{\theta}})
= -\mathbb{E}_{(\boldsymbol{x}, \tau) \sim \mathcal{D}}
\big[ \log p_S^{\boldsymbol{\theta}}(\tau \mid \boldsymbol{x}) \big] \).

\subsection{Progress Stagnation in SFT-Distilled Agents}\label{sec:progress_stagnation}
While SFT distillation enables SLMs to develop proficiency in tool-augmented reasoning and reduces basic tool-invocation errors, our analysis identifies a fundamental limitation: SLM agents frequently exhibit \textit{progress stagnation} when tackling complex tasks. As the interaction context grows with extensive reasoning chains and environmental feedback, such as noisy search results and verbose code-execution logs, these agents frequently lapse into unproductive reasoning-action loops. In these scenarios, the student model tends to redundantly recycle ineffective strategies rather than synthesize feedback to generate novel solution directions, thereby impeding task progress. As illustrated in \cref{fig:case} (right), an SFT-distilled agent tasked with a factual reasoning query repeatedly issues highly similar web-search actions across multiple turns, despite prior attempts failing to retrieve the requisite information. This persistent failure to adapt eventually leads to task expiration or incorrect termination. In contrast, the teacher agent maintains steady progress (\cref{fig:case}, left): by effectively leveraging prior interactions, the teacher adaptively refines its actions, progressively acquiring new information until the task is successfully resolved.

\begin{wrapfigure}{r}{0.4\textwidth}
\vspace{-33pt}
  \begin{minipage}{\linewidth}
\begin{table}[H]
\caption{Three formulations of the turn-level stagnation score
\(S_{\boldsymbol{h}_i \oplus \boldsymbol{a}_i}\).}
\label{tab:three_metrics}
\vspace{0.4em}
\centering
\resizebox{1.0\linewidth}{!}{
\begin{tabular}{l c}
\toprule
\textbf{Score} &  \textbf{Formulation} \\
\midrule
Lexical  & $\max_{j < i} \text{ROUGE}_{\text{L}}(\boldsymbol{h}_i \oplus \boldsymbol{a}_i,\boldsymbol{h}_j \oplus \boldsymbol{a}_j)$ \\ [0.3em]
Semantic & $\max_{j < i} \text{Cos}(E(\boldsymbol{h}_i \oplus \boldsymbol{a}_i),E(\boldsymbol{h}_j \oplus \boldsymbol{a}_j))$ \\ [0.3em]
Judgment  & $J_{\text{LLM}}(\boldsymbol{h}_i \oplus \boldsymbol{a}_i \mid \{\boldsymbol{h}_j \oplus \boldsymbol{a}_j\}_{j<i})$ \\
\bottomrule
\end{tabular}}
\end{table}
\end{minipage}
\vskip -0.18in
\end{wrapfigure}
\textbf{Stagnation Score.} To quantify the gap between student and teacher agents in mitigating progress stagnation, we operationalize the turn-level \textit{stagnation
score} in three variants: lexical overlap, semantic similarity, and LLM-as-judge evaluation. Specifically, for the thought
\(\boldsymbol{h}_i\) and action \(\boldsymbol{a}_i\) generated at turn \(i\), let \(\boldsymbol{h}_i \oplus \boldsymbol{a}_i\) denote their concatenation. The three formulations of the stagnation score
$S_{\boldsymbol{h}_i \oplus \boldsymbol{a}_i} \in [0,1]$ are defined in~\cref{tab:three_metrics}, and all target the same underlying phenomenon — whether the current turn revisits previously attempted thoughts or actions. Lexical overlap, measured by ROUGE-L~\citep{rouge}, provides a clear and interpretable signal, particularly informative in agent settings where actions often involve code generation or API calls rather than free-form text. Semantic similarity captures stagnation beyond surface-form overlap by detecting paraphrased recurrences. We encode each thought-action pair into a sentence embedding \(E(\boldsymbol{h}_i \oplus \boldsymbol{a}_i)\) using all-mpnet-base-v2 \citep{sentencebert} and compute the cosine similarity between embedded pairs.  LLM-as-judge employs a frontier LLM to directly assess turn-level progress following the protocol of \citet{oagent} (see Appendix~\ref{app:judgment} for details). For lexical and semantic scores, we compare the current turn against all previous turns and take the maximum, rather than comparing only with the immediate predecessor, since stagnation loops can span multiple turns. The LLM-as-judge score implicitly accounts for the full prior history through its prompt design. Since different tasks may naturally exhibit varying degrees of benign overlap between interaction turns, we primarily analyze the teacher--student gap in stagnation scores rather than absolute values. This gap measures the behavioral discrepancy in maintaining progress and avoiding unproductive cycles.
\begin{figure}[t]
\begin{center}
\centerline{\includegraphics[width=1.0\linewidth]{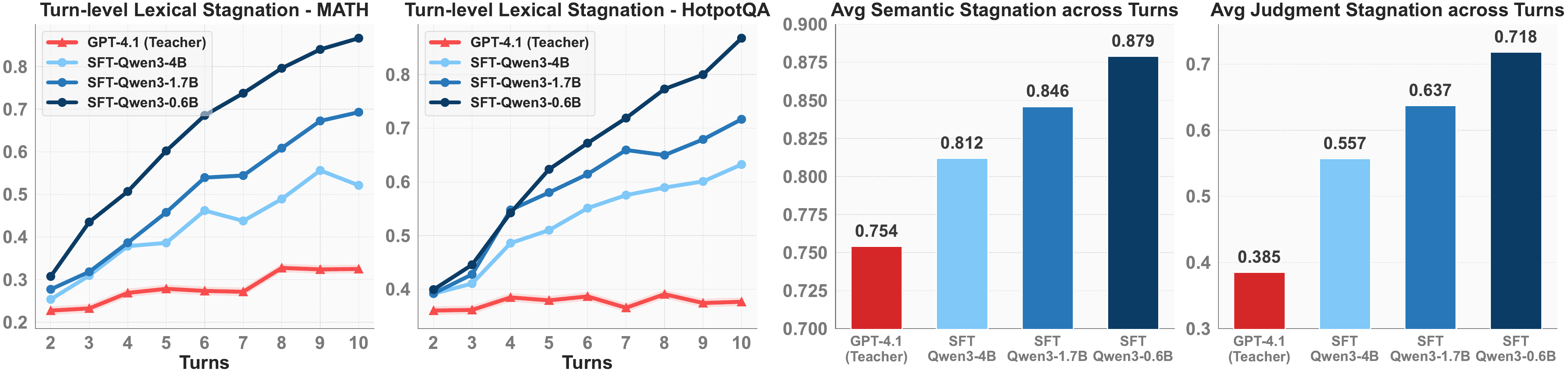}}
\vskip -0.1in
\caption{
Comparison of stagnation scores between the teacher agent (GPT-4.1) and SFT-distilled students (Qwen3-4/1.7/0.6B): \textbf{(1\&2)} turn-level lexical stagnation on MATH and HotpotQA, \textbf{(3\&4)} average semantic and judgment stagnation across turns and datasets.}
\label{fig:progress_stagnation}
\end{center}
\vskip -0.4in
\end{figure}

\begin{wrapfigure}{r}{0.39\textwidth}
\vspace{-33pt}
  \begin{minipage}{\linewidth}
\begin{table}[H]
\caption{Pearson correlation among stagnation metrics on Qwen SFT agents, averaged across HotpotQA and MATH.}
\label{tab:pearson}
\vspace{0.3em}
\centering
\resizebox{1.0\linewidth}{!}{
\begin{tabular}{lccc}
\toprule
\textbf{Pearson \(r\)} 
& \textbf{4B} 
& \textbf{1.7B} 
& \textbf{0.6B} \\
\midrule
$r(S^{\mathrm{lex}}, S^{\mathrm{sem}})$     & 0.929 & 0.958 & 0.964 \\
$r(S^{\mathrm{lex}}, S^{\mathrm{judge}})$   & 0.946 & 0.975 & 0.956 \\
$r(S^{\mathrm{sem}}, S^{\mathrm{judge}})$   & 0.947 & 0.971 & 0.979 \\
\bottomrule
\end{tabular}}
\end{table}
\end{minipage}
\vskip -0.125in
\end{wrapfigure}
\textbf{Stagnation Evaluation.} Equipped with the proposed stagnation metric, we conduct a comprehensive analysis under the CodeAct framework across diverse agentic tasks. Following the SFT distillation procedure in \cref{sec:preliminary}, we employ GPT-4.1 as the teacher agent to collect 3,000 trajectories from the training sets of MATH \citep{MATH} and HotpotQA \citep{hotpotqa}, respectively. With appropriate tool set configurations \citep{agentdistill}, these benchmarks facilitate the evaluation of diverse agentic capabilities, including code-assisted numerical computation, software package execution, and web information seeking. We then distill these capabilities into various SLMs using LoRA \citep{lora}, with training details provided in Appendix \ref{app:hyper}. \cref{fig:progress_stagnation} (1\&2) presents turn-level lexical stagnation scores for distilled Qwen3-4B/1.7B/0.6B on held-out evaluation sets, while \cref{fig:progress_stagnation} (3\&4) report the average semantic and judgment stagnation across turns and datasets. The results reveal a consistent pattern: distilled SLMs exhibit substantially more severe stagnation than the teacher agent, and this disparity becomes increasingly pronounced as model scale decreases or trajectory length increases. In particular, the lexical stagnation score can reach values close to 0.9, indicating that the student actions highly resemble their previous attempts. We further compute pairwise Pearson correlations \(r\) among the three stagnation metrics in \cref{tab:pearson}, finding consistently high agreement with \(r > 0.92\). This strong correlation suggests that, under severe stagnation, the three metrics capture the same underlying failure mode in SLM agents. We observe the similar stagnation trend across additional model families, including Llama3 and Gemma3 (see \cref{sec:main_result} and Appendix \ref{app:gemma}). The pattern also persists under full-parameter fine-tuning, as detailed in Appendix \ref{app:full_fine_tune}, and remains evident even when evaluating on the distillation training data itself (Appendix \ref{app:stagnation_training_set}). These findings collectively demonstrate that SFT distillation alone is insufficient for enabling SLMs to acquire the teacher's resilience to stagnation.

\section{Progress-Aware Distillation for SLM Agents} \label{sec:ProDi}
Building on our analysis in \cref{sec:stagnation_analysis}, we observe that standard SFT distillation often leads to suboptimal outcomes since student agents fail to replicate the teacher’s non-stagnant progression. Motivated by this gap, we propose \textbf{Pro}gress-Aware \textbf{D}istillation (\textbf{ProD}), a training strategy that explicitly detects stagnation in student-generated trajectories and applies adaptive penalties during optimization. By discouraging unproductive reasoning-action patterns, ProD enables SLM agents to maintain forward progress and more closely align with the teacher’s problem-solving dynamics.

\textbf{Penalizing Stagnation in Student Trajectories.} Our approach starts with a base student agent  $p_S^{\boldsymbol{\theta}_0}$ obtained via SFT. Then, for each query $\boldsymbol{x}$ from the SFT distillation dataset $\mathcal{D}$, we execute the student agent to generate one trajectory $
\tilde{\tau} = \big[
(\tilde{\boldsymbol{h}}_{1}, \tilde{\boldsymbol{a}}_{1}, \tilde{\boldsymbol{o}}_{1}),\ldots,
(\tilde{\boldsymbol{h}}_{L_{\tilde{\tau}}}, \tilde{\boldsymbol{a}}_{L_{\tilde{\tau}}}, \tilde{\boldsymbol{o}}_{L_{\tilde{\tau}}})
\big]
\sim p_S^{\boldsymbol{\theta}_0}(\cdot \mid \boldsymbol{x})
$. Our goal is to favor the teacher’s forward-progressing trajectories while penalizing stagnant behavior in student trajectories. A central challenge is modulating the penalty strength: student trajectories with severe stagnation require stronger corrective signals, whereas those with smoother progress should receive milder updates to avoid overcorrection. To this end, we encode the discrepancy in stagnation levels between the student and teacher into an adaptive margin $f(\tilde{\tau},\tau)$ (defined later), which leads to the following progress-aware objective:
\begin{equation}
\label{eq:prodi}
\scalebox{0.95}{$
\mathcal{L}_{\text{PRO}}(\boldsymbol{\theta}) =
-\mathbb{E}_{(\boldsymbol{x}, \tau)\sim\mathcal{D},\,
                 \tilde{\tau}\sim p_S^{\boldsymbol{\theta}_0}(\cdot\mid \boldsymbol{x})}
\Big[
    \log \sigma \Big(
        \frac{\beta}{\lvert \tau\rvert}\,
        \log p_S^{\boldsymbol{\theta}}(\tau \mid \boldsymbol{x})
        -\frac{\beta}{\lvert \tilde{\tau}\rvert}\,
         \log p_S^{\boldsymbol{\theta}}(\tilde{\tau} \mid \boldsymbol{x})
         - f(\tilde{\tau},\tau)
    \Big)
\Big],
$}
\end{equation}
where $\sigma$ is the sigmoid function,
$\lvert\tau\rvert$ and $\lvert\tilde{\tau}\rvert$ denote the token lengths of teacher and student trajectories, and $\beta$ scales the log-likelihood gap. This formulation adaptively scales the gradient signal, whereby student trajectories with pronounced stagnation yield larger margins and induce aggressive corrective updates, while trajectories approaching the teacher's progress dynamics receive smaller margins and are preserved with minimal intervention. A detailed gradient analysis is provided in Appendix \ref{app:gradient}.

\textbf{Adaptive Margin.} We first define a trajectory-level stagnation gap as the difference between the student’s and teacher’s average turn-wise stagnation:
\begin{equation}
\label{eq:stagnation_gap}
\scalebox{1.0}{$
\Delta S(\tilde{\tau},\tau)
:= 
\frac{1}{L_{\tilde{\tau}}}\sum_{i=1}^{L_{\tilde{\tau}}} S_{\boldsymbol{\tilde{h}}_i \oplus \boldsymbol{\tilde{a}}_i}
-
\frac{1}{L_\tau}\sum_{i=1}^{L_\tau} S_{\boldsymbol{h}_i \oplus \boldsymbol{a}_i},
$}
\end{equation}
where the stagnation score for the first turn is set to zero, since there are no prior turns to compare against. Using a base margin constant $c_0$, we construct an adaptive margin as
$c_0 \cdot \max\{0,\Delta S(\tilde{\tau},\tau)\}$, which increases proportionally to the severity of the student's stagnation relative to the teacher. To further distinguish student trajectories that yield correct answers from those that do not, we incorporate answer correctness into the margin via a weighted combination:
\begin{equation}
\label{eq:margin}
\scalebox{0.92}{$
f(\tilde{\tau},\tau)
=
c_0 \Big(
\alpha \max\{0,\Delta S(\tilde{\tau},\tau)\}
+
(1-\alpha)\mathbb{I}_{\text{Ans}(\tilde{\tau}) \neq \text{Ans}(\tau)}
\Big),
$}
\end{equation}
where $\text{Ans}(\cdot)$ extracts the final answer from a trajectory and $\alpha \in [0,1]$ balances the contributions from stagnation severity and answer correctness. Under this formulation, student trajectories that simultaneously exhibit pronounced stagnation and produce incorrect answers receive the largest margins, thereby inducing stronger corrective updates.

\begin{wrapfigure}{r}{0.54\textwidth}
\vspace{-2.3em}
\begin{minipage}{0.54\textwidth}
\begin{algorithm}[H]
\small
\caption{Progress-Aware Distillation (ProD)} \label{alg:prodi}
\textbf{Input:} SFT dataset \(\mathcal{D}=\{(\boldsymbol{x}_i,\tau_i)\}_{i=1}^N\), SFT student \(p_S^{\boldsymbol{\theta}_0}\), and iterations \(T\)
\begin{algorithmic}[1]
\FOR{$t=0$ {\bfseries to} $T-1$}
\FOR{$i=1$ {\bfseries to} $N$}
\STATE Generate a student trajectory $\tilde{\tau}_i \sim p_S^{\boldsymbol{\theta}_t}(\cdot \mid \boldsymbol{x}_i)$
\STATE Compute stagnation gap $\Delta S(\tilde{\tau}_i, \tau_i)$ using Eq.~(\ref{eq:stagnation_gap})
\STATE Compute adaptive margin $f(\tilde{\tau}_i, \tau_i)$ using Eq.~(\ref{eq:margin})
\ENDFOR
\STATE 
   $
    \mathcal{L}({\boldsymbol{\theta}})
    = \frac{1}{N}
     \sum_{i=1}^N 
    \Big[
        -\log \sigma\!\Big(
        \frac{\beta}{\lvert\tau_i\rvert}\log p_S^{\boldsymbol{\theta}}(\tau_i \mid \boldsymbol{x}_i)
        -\frac{\beta}{\lvert\tilde{\tau}_i\rvert}\log p_S^{\boldsymbol{\theta}}(\tilde{\tau}_i \mid \boldsymbol{x}_i)
        - f(\tilde{\tau}_i,\tau_i)
        \Big)
        - \lambda \log p_S^{\boldsymbol{\theta}}(\tau_i \mid \boldsymbol{x}_i)
    \Big]
    $
\STATE Update $\boldsymbol{\theta}_{t+1} \gets \arg\min_{\boldsymbol{\theta}} \mathcal{L}({\boldsymbol{\theta}})$
\ENDFOR\\
\end{algorithmic}
\textbf{Output:} $\boldsymbol{\theta}_T$
\end{algorithm}
\end{minipage}
\vspace{-1em}
\end{wrapfigure}
\textbf{Overall Optimization Objective.} The objective in \cref{eq:prodi} is inspired by SimPO \citep{simpo}. However, unlike SimPO, which relies on offline preference data with a fixed margin shared across instances, our formulation leverages student generations from the current model and introduces an instance-dependent margin that more accurately reflects the behavioral discrepancies between the student and teacher. While $\mathcal{L}_{\text{PRO}}$ effectively widens the relative probability gap, optimizing it in isolation may trigger likelihood displacement \citep{likelihood-displace}, where the probabilities of both teacher and student trajectories decrease simultaneously. This is undesirable because the teacher trajectory encodes the target problem-solving behavior that the student should preserve. To mitigate this, we retain the original SFT loss and jointly optimize it with the progress-aware objective, \(
\mathcal{L}({\boldsymbol{\theta}}) =   \mathcal{L}_{\text{PRO}}({\boldsymbol{\theta}}) + \lambda \,\mathcal{L}_{\text{SFT}}({\boldsymbol{\theta}})
\), thereby maintaining the likelihood of high-quality teacher trajectories while still applying effective corrective signals to stagnant student trajectories. In practice, if a student agent still demonstrates a substantial stagnation gap relative to the teacher after a single optimization round, the distillation process can be executed iteratively. In each iteration, the student agent from the previous round is used to generate fresh trajectories and recompute stagnation penalties, thereby progressively nudging the student toward the teacher’s behavior. The overall ProD procedure is summarized in \cref{alg:prodi}.

\begin{table*}[t]
  \caption{Comparison between teacher and student agents, including base models, SFT distillation, and ProD, on in-domain and out-of-domain agent tasks, evaluated by final answer accuracy (\%).}
  \vskip -0.1in
  \label{tab:main_result}
  \begin{center}
    \resizebox{0.95\linewidth}{!}{
        \begin{tabular}{l c c c c c c c}
          \toprule
          \multirow{3}{*}{\raisebox{-0.8ex}{\textbf{Method}}}  & \multicolumn{3}{c}{\textbf{In-domain}}  & \multicolumn{4}{c}{\textbf{Out-of-domain}}   \\
          \cmidrule(lr){2-4} \cmidrule(lr){5-8}
& \textbf{MATH}
& \textbf{HotpotQA}
& \textbf{Avg.}
& \makecell[c]{\textbf{Omni-MATH}\\\textbf{(Level 1--3)}}
& \makecell[c]{\textbf{Omni-MATH}\\\textbf{(Level 4--5)}}
& \makecell[c]{\textbf{2Wiki}\\\textbf{MultiHopQA}}
& \textbf{Avg.} \\
          \midrule
\rowcolor{gray!20}\multicolumn{8}{c}{\textit{\textbf{Teacher Agent}}} \\
GPT-4.1 & 90.4 & 71.0 & 80.7 & 82.0 & 60.0 & 59.2 & 67.1\\
\midrule
\rowcolor{gray!20}\multicolumn{8}{c}{\textit{\textbf{Student Agents}}} \\
Qwen3-4B & 56.8 & 53.8 & 55.3 & 57.8 & 23.0 & 44.8 & 41.9\\
SFT Distill & 81.9 & 66.6 & 74.2 & 73.6 & 39.4 & 49.4 & 54.1\\
\textbf{ProD} & \textbf{85.2} & \textbf{70.8} & \textbf{78.0} & \textbf{76.8} & \textbf{46.0} & \textbf{52.4} & \textbf{58.4}\\
\midrule
Qwen3-1.7B & 46.1 & 41.6 & 43.8 & 48.4 & 13.8 & 29.2 & 30.5\\
SFT Distill & 65.7 & 58.6 & 62.2 & 59.2 & 23.0 & 42.8 & 41.7\\
\textbf{ProD} & \textbf{70.8} & \textbf{65.0} & \textbf{67.9} & \textbf{64.8} & \textbf{28.2} & \textbf{51.2} & \textbf{48.1}\\
\midrule
Qwen3-0.6B & 17.8 & 3.4 & 10.6 & 25.4 & 3.6 & 0.0 & 9.7\\
SFT Distill & 46.0 & 52.0 & 49.0 & 46.8 & 10.4 & 34.2 & 30.5\\
\textbf{ProD} & \textbf{50.0} & \textbf{55.4} & \textbf{52.7} & \textbf{51.6} & \textbf{14.6} & \textbf{38.4} & \textbf{34.9}\\
\midrule
Llama-3.1-8B-Inst & 50.1 & 45.8 & 48.0 & 47.8 & 14.6  & 37.4 & 33.3\\
SFT Distill & 59.1 & 64.6 & 61.8 & 52.8 & 12.6 & 51.8 & 39.1\\
\textbf{ProD} & \textbf{65.2} & \textbf{69.2} & \textbf{67.2} & \textbf{59.8} & \textbf{18.6} & \textbf{57.6} & \textbf{45.3}\\
\midrule
Llama-3.2-3B-Inst & 27.8 & 19.8 & 23.8 & 24.8 & 5.8 & 15.4 & 15.3\\
SFT Distill & 55.5 & 61.6 & 58.6 & 49.0 & 11.0 & 45.4 & 35.1\\
\textbf{ProD} & \textbf{58.6} & \textbf{67.0} & \textbf{62.8} & \textbf{53.6} & \textbf{14.8} & \textbf{51.2} & \textbf{39.9}\\
\midrule
Llama-3.2-1B-Inst & 2.6 & 6.0 & 4.3 & 2.6 & 2.0 & 0.0 & 1.5\\
SFT Distill & 25.6 & 46.2 & 35.9 & 27.8 & \textbf{4.8} & 28.0 & 20.2\\
\textbf{ProD} & \textbf{30.3} & \textbf{52.8} & \textbf{41.6} & \textbf{33.6} & \textbf{4.8} & \textbf{38.6} & \textbf{25.7}\\
\midrule
Gemma-3-4b-it & 42.0 & 41.2 & 41.6 & 43.8 & \textbf{15.2} & 39.6 & 32.9\\
SFT Distill & 50.5 & 62.2 & 56.4 & 53.2 & 14.8 & 44.6 & 37.5\\
\textbf{ProD} & \textbf{55.6} & \textbf{66.4} & \textbf{61.0} & \textbf{55.6} & 15.0 & \textbf{49.4} & \textbf{40.0}\\
\midrule
Gemma-3-1b-it & 3.1 & 4.6 & 3.8 & 2.8 & 0.2 & 1.0 & 1.3\\
SFT Distill & 16.0 & 41.2 & 28.6 & 22.2 & 1.4 & 25.8 & 16.5\\
\textbf{ProD} & \textbf{19.6} & \textbf{46.8} & \textbf{33.2} & \textbf{26.2} & \textbf{3.4} & \textbf{36.0} & \textbf{21.9}\\
          \bottomrule
        \end{tabular}}
  \end{center}
  \vskip -0.25in
\end{table*}
\section{Experiments} \label{sec:exp}
\subsection{Experimental Settings}
\textbf{Student Backbones and Teacher Agent.} To evaluate the effectiveness of ProD across various model scales, we select eight SLMs ranging from 0.6B to 8B parameters as student backbones. Specifically, our evaluation includes the Qwen3 series (4B, 1.7B, 0.6B) \citep{qwen3}, Llama-3.1-8B-Instruct \citep{llama3}, Llama-3.2 (3B, 1B-Instruct) \citep{llama3}, and Gemma-3 (4B, 1B-it) \citep{gemma3}. Following the SFT distillation procedure described in \cref{sec:preliminary}, we employ GPT-4.1 as the teacher agent to collect 3,000 trajectories from the training sets of MATH \citep{MATH} and HotpotQA \citep{hotpotqa}, respectively, and retain only the successful ones to construct the distillation dataset.

\textbf{Implementation Details.} All experiments use the CodeAct framework \citep{codeact}, with the web-search tool implemented following \citep{agentdistill,search-r1}. For both SFT and ProD, we train separate LoRA weights \citep{lora} for each task with rank 64, batch size 32, AdamW, and a linear learning rate schedule with 3\% warmup. ProD initializes from the SFT-distilled student agent and is further optimized. The main results in \cref{sec:main_result} use the lexical stagnation score, a single ProD iteration, and an adaptive-margin weighting of \(\alpha=0.5\). We analyze iterative ProD optimization in \cref{sec:more_iter}, and report results using alternative stagnation scores in Appendix \ref{app:alternative}. During inference, we use greedy decoding and cap the interaction horizon at 10 turns. Additional hyperparameter details (e.g., learning rate, training epochs, $c_0$, and $\lambda$) are provided in Appendix~\ref{app:hyper}.

\textbf{Datasets and Evaluation.} We evaluate agentic capabilities primarily through factual and mathematical reasoning tasks, covering code-assisted numerical computation, software execution, and web-based information seeking. We consider both in-domain (ID) and out-of-domain (OOD) generalization. For ID evaluation, we use MATH and HotpotQA. We sample 110 problems from each of the seven MATH subjects (770 in total) and 500 hard-level questions from HotpotQA. To assess OOD generalization, we utilize Omni-MATH \citep{omnimath}, sampling 500 problems from difficulty levels 1–3 and another 500 from levels 4–5. We also include 500 instances from 2WikiMultiHopQA \citep{2wiki} to provide broader coverage of web-search-based multi-hop factual reasoning. Following prior work \citep{agentdistill, omnimath}, we rely on an LLM-as-Judge (GPT-4.1-mini) to determine accuracy by comparing generated answers with ground-truth reference. Evaluation prompts are provided in Appendix \ref{app:eval_prompt}. Additionally, we use stagnation scores and related efficiency metrics to quantify the practical utility of our method in reducing stagnant interaction loops.

\begin{figure}[t]
\begin{center}
\centerline{\includegraphics[width=1.0\linewidth]{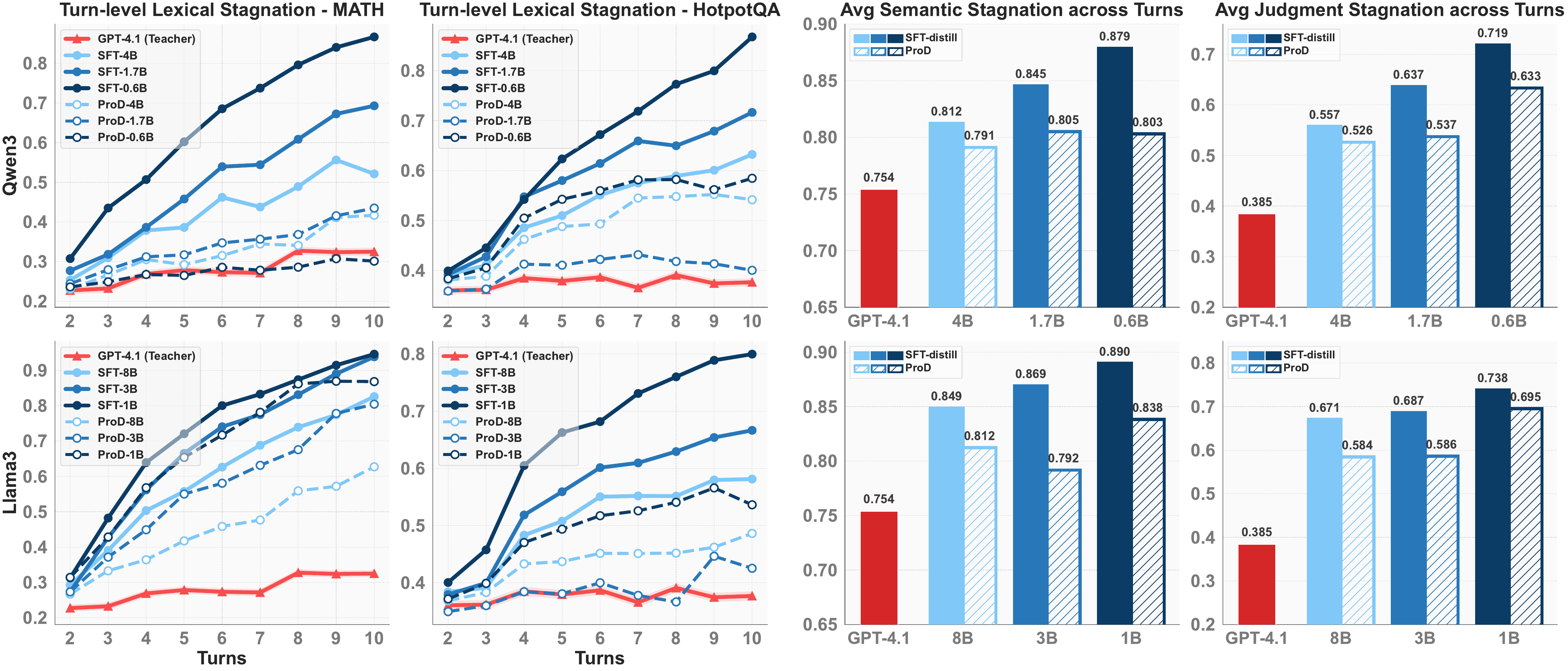}}
\vskip -0.1in
\caption{
Alleviating progress stagnation.
Comparison between SFT and ProD agents:
\textbf{Columns 1\&2} show turn-level lexical stagnation on MATH and HotpotQA,
\textbf{Columns 3\&4} show average semantic and judgment stagnation across turns and datasets.
}
\label{fig:mitigate_stagnation}
\end{center}
\vskip -0.4in
\end{figure}

\vspace{-0.5em}
\subsection{Main Results} \label{sec:main_result}
\textbf{ProD consistently surpasses SFT distillation in both in-domain and out-of-domain settings.} \cref{tab:main_result} reports the accuracy of 8 student agents across multiple benchmarks. Compared with the base models, ProD yields substantial absolute improvements and consistently outperforms standard SFT distillation across all model families. On average, ProD improves in-domain accuracy by 29.2\% over base models and by 4.7\% over SFT, while achieving out-of-domain gains of 18.5\% and 4.9\%, respectively. Notably, ProD exhibits average relative improvement of approximately 10\% in-domain and 17\% out-of-domain over SFT. ProD enables students to narrow the gap to their teacher. For instance, on HotpotQA, the Qwen3-4B student distilled via ProD achieves 70.8\% accuracy, nearly matching the teacher's performance of 71.0\%. ProD also outperforms rejection-sampling SFT, as shown in Appendix~\ref{app:rs_sft}. These results underscore ProD's ability to enhance both the fundamental effectiveness and the generalization of SLM-based agents.

\begin{wrapfigure}{r}{0.34\textwidth}
\vspace{-32pt}
  \begin{minipage}{\linewidth}
\begin{table}[H]
\caption{Inference efficiency averaged over 8 student agents on ID and OOD settings.}
\label{tab:efficiency}
\vspace{0.4em}
\centering
\resizebox{1.0\linewidth}{!}{
\begin{tabular}{lcc}
\toprule
\textbf{Cost} & \textbf{SFT} & \textbf{ProD} \\
\midrule
Avg Turns  & 4.144   & 3.663 ($\downarrow$11.6\%) \\
Avg Tokens & 16206 & 12847 ($\downarrow$20.7\%) \\
\bottomrule
\end{tabular}}
\end{table}
\end{minipage}
\vskip -0.13in
\end{wrapfigure}
\textbf{ProD effectively mitigates progress stagnation and enhances task efficiency.}
In \cref{fig:mitigate_stagnation}, we compare the stagnation patterns of SFT and ProD agents. The results show that ProD substantially alleviates stagnation across lexical, semantic, and LLM-as-judge metrics, allowing student agents to maintain smoother trajectories that align more closely with the teacher's behavior. Additional Gemma3 results are provided in Appendix \ref{app:gemma}, and qualitative examples are shown in Appendix~\ref{app:our_case}. Beyond trajectory quality, \cref{tab:efficiency} reports the average number of interaction turns and token consumption, averaged over eight student agents in both ID and OOD settings. By explicitly penalizing stagnant reasoning-action patterns, ProD enables agents to solve tasks more decisively. This leads to a consistent reduction in interaction length across most SLMs on all benchmarks, demonstrating that ProD not only improves accuracy but also significantly enhances overall task efficiency. We defer the analysis of training overhead to Appendix \ref{app:train_cost}.

\subsection{Comparison with Repetition-Control Methods}
In this section, we investigate whether existing repetition-control techniques can help address progress stagnation in SLM agents. We compare against two representative methods that suppress redundant generation at the decoding level and at the training-data level, respectively. Specifically, Repetition Penalty (RP)~\citep{keskar2019ctrl,vllm_repetition_penalty} operates at decoding time by rescaling the logits of previously generated tokens to discourage excessive repetition. We apply RP on top of SFT-distill with penalty strengths of \(1.1\), \(1.2\), and \(1.3\). The other family reduces redundancy at the data level by shortening teacher demonstrations before distillation. For example, \citet{concisedistill-0} compress long chain-of-thought traces into concise variants. In the agent-distillation setting, we implement this strategy by identifying training examples whose teacher trajectory contains at least 4 steps and prompting the teacher to re-execute the task with its original trajectory as context, producing shorter and less redundant demonstrations (see Appendix~\ref{app:concise} for the detailed prompt). This procedure reduces the average number of steps from \(4.98\) to \(2.65\). As shown in Table~\ref{tab:repcontrol}, ProD substantially outperforms both methods, indicating that decoding-level diversity enforcement is insufficient to address trajectory-level stagnation, while overly large penalties further degrade performance. Data-level shortening reduces redundancy in the teacher demonstrations, but does not impose any constraint on the student's own behavior during optimization, yielding only minimal gains over SFT distillation.
\begin{table}[h]
\vskip -0.15in
\centering
\caption{Comparison with repetition-control baselines on HotpotQA. RP$(\cdot)$ denotes repetition penalty with the corresponding strength. Concise SFT trains on shortened, less redundant teacher trajectories.}
\label{tab:repcontrol}
\vskip 0.05in
\resizebox{0.7\linewidth}{!}{
\begin{tabular}{lcccccc}
\toprule
\textbf{Model} & \textbf{SFT} & \textbf{RP(1.1)} & \textbf{RP(1.2)} & \textbf{RP(1.3)} & \textbf{Concise SFT} & \textbf{ProD} \\
\midrule
Qwen3-4B     & 66.6 & 66.2 & 65.6 & 61.6 & 66.4 & \textbf{70.8} \\
Llama3.2-3B  & 61.6 & 62.0 & 61.2 & 58.0 & 62.0 & \textbf{67.0} \\
\bottomrule
\end{tabular}}
\vskip -0.2in
\end{table}

\begin{figure}[t]
\begin{center}
\centerline{\includegraphics[width=0.98\linewidth]{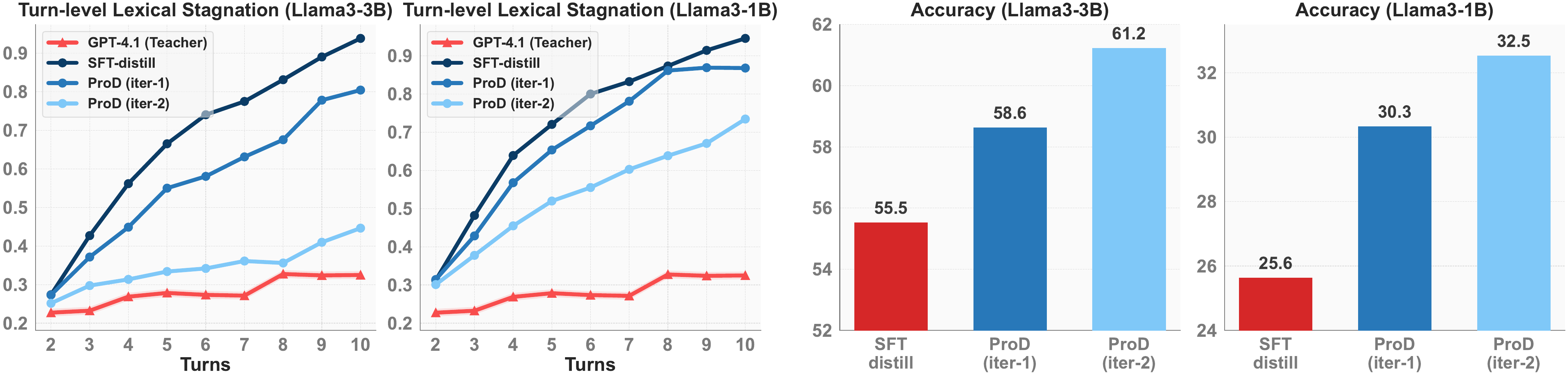}}
\vskip -0.1in
\caption{Iterative ProD further alleviates stagnation \textbf{(1\&2)} and improves accuracy \textbf{(3\&4)} on MATH.}
\label{fig:more_iter}
\end{center}
\vskip -0.35in
\end{figure}

\subsection{Effects of Iterative ProD Optimization} \label{sec:more_iter}
\cref{fig:mitigate_stagnation} shows that even a single ProD iteration substantially improves distilled agents' resistance to progress stagnation.  For the Qwen family, ProD students achieve stagnation scores close to those of the teacher. However, for other model families, such as Llama3.2-3B and Llama3.2-1B on MATH, a noticeable gap remains despite improvements over standard SFT.  Motivated by this, we study additional ProD iterations by using the first-iteration students as new base models, regenerating trajectories, and recomputing stagnation gaps and adaptive margins for further distillation. \cref{fig:more_iter} (1\&2) illustrates the stagnation scores after two ProD iterations. The results indicate that two ProD iterations further reduce stagnation, pushing student behavior closer to the teacher's progress dynamics. This refinement also yields consistent gains in final answer accuracy, as shown in \cref{fig:more_iter} (3\&4). These results suggest that iterative ProD can further alleviate persistent stagnation and improve performance when a single round is insufficient to close the teacher--student behavioral gap.

\subsection{Ablation Studies} \label{sec:ablation}
In this section, we examine the key components of ProD, including the adaptive margin and the SFT loss term. We further analyze the sensitivity of the adaptive margin by varying the relative weighting between the stagnation gap and final answer correctness. The results validate these design choices and indicate that both signals are necessary to capture behavioral discrepancies between the student and teacher. Hyperparameter details for the ablations are provided in Appendix \ref{app:hyper}.

\begin{figure}[t]
\begin{center}
\centerline{\includegraphics[width=1.0\linewidth]{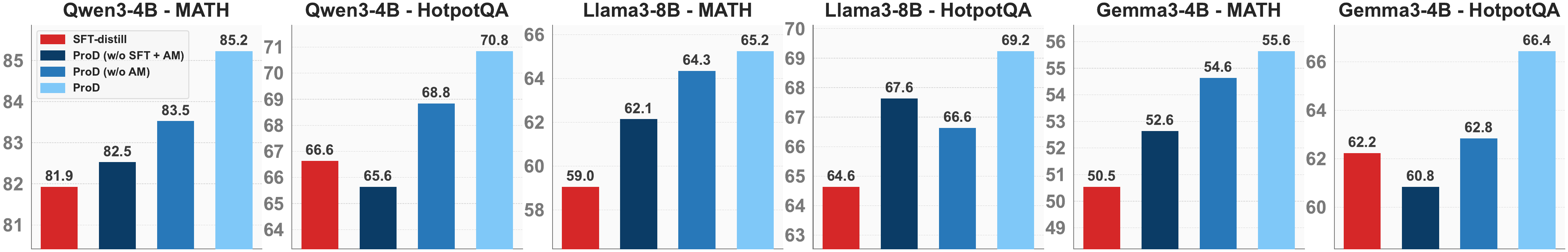}}
\vskip -0.05in
\caption{Ablation of optimization objective components in ProD with two ablated variants, ProD (w/o SFT + AM) and ProD (w/o AM), where AM denotes the adaptive margin.}
\label{fig:ablation_component}
\end{center}
\vskip -0.39in
\end{figure}

\textbf{Ablation of Optimization Components.} We evaluate two variants to assess the roles of the adaptive margin and the SFT objective: (1) \textbf{ProD (w/o SFT + AM)} removes the SFT loss from \(
\mathcal{L}({\boldsymbol{\theta}}) \) and replaces the adaptive margin with a fixed constant. This version effectively degenerates to SimPO \citep{simpo}, albeit using trajectories produced by the current student agent. (2) \textbf{ProD (w/o AM)} retains the SFT loss but adopts a fixed margin. As shown in \cref{fig:ablation_component}, ProD (w/o SFT + AM) exhibits unstable performance and even underperforms the SFT-distilled baseline on Qwen3-4B and Gemma3-4B-it. This underscores the importance of the SFT loss in anchoring the student to valid reasoning paths by maintaining high probability on teacher trajectories. In contrast, ProD (w/o AM) consistently outperforms SFT but remains inferior to the full ProD formulation. This suggests that a fixed margin may lead to overcorrection or insufficient optimization for varied trajectory qualities, thereby limiting potential gains. These results indicate that the SFT loss and adaptive margin play complementary roles in stabilizing optimization and bolstering performance.

\begin{wrapfigure}{r}{0.43\textwidth}
\vspace{-33pt}
  \begin{minipage}{\linewidth}
\begin{table}[H]
  \caption{Accuracy (\%) under different \(\alpha\) values in the adaptive margin.}
\label{tab:ablation_alpha}
\vspace{0.4em}
\centering
\resizebox{1.0\linewidth}{!}{
        \begin{tabular}{c c c c c c c}
          \toprule
          \multirow{2}{*}{\raisebox{-0.7ex}{\textbf{Dataset}}} & \multirow{2}{*}{\raisebox{-0.7ex}{\textbf{SFT}}}  & \multicolumn{5}{c}{\textbf{ProD}} \\
          \cmidrule(lr){3-7}
          & & $\alpha=1$ & $\alpha=0.75$ & $\alpha=0.5$ & $\alpha=0.25$ & $\alpha=0$  \\
\midrule
        \rowcolor{gray!20}\multicolumn{7}{c}{\textit{\textbf{Qwen3-4B}}} \\
        \textbf{MATH} & 81.9 & 85.2 & \textbf{85.8} & 85.2 & 84.5 & 85.3\\
        \textbf{HotpotQA} & 66.6 & 69.4 & 68.8 & \textbf{70.8} & 66.8 & 67.6\\
        \textbf{Average} & 74.2 & 77.3 & 77.3 & \textbf{78.0} & 75.7 & 76.4\\
\midrule
        \rowcolor{gray!20}\multicolumn{7}{c}{\textit{\textbf{Llama3-8B}}} \\
        \textbf{MATH} & 59.1 & 64.5 & 65.1 & \textbf{65.2} & 64.2 & 65.1\\
        \textbf{HotpotQA} & 64.6 & 67.8 & 67.8 & \textbf{69.2} & 66.6 & 67.4\\
        \textbf{Average} & 61.8 & 66.2 & 66.4 & \textbf{67.2} & 65.4 & 66.2\\
\midrule
        \rowcolor{gray!20}\multicolumn{7}{c}{\textit{\textbf{Gemma3-4B}}} \\
        \textbf{MATH} & 50.5 & 55.2 & 54.7 & \textbf{55.6} & 54.0 & 54.3\\
        \textbf{HotpotQA} & 62.2 & 65.8 & 65.2 & \textbf{66.4} & 64.6 & 64.8\\
        \textbf{Average} & 56.4 & 60.5 & 60.0 & \textbf{61.0} & 59.3 & 59.6\\
          \bottomrule
        \end{tabular}}
\end{table}
\end{minipage}
\vskip -0.2in
\end{wrapfigure}
\textbf{Effect of Adaptive Margin Weighting.} The coefficient \(\alpha\) in \cref{eq:margin} balances the stagnation gap and answer correctness, with \(\alpha=0.5\) used by default in \cref{tab:main_result}. 
\cref{tab:ablation_alpha} reports ProD performance under different \(\alpha\) values, where \(\alpha=1\) uses only the stagnation gap and \(\alpha=0\) uses only answer correctness. Even with the stagnation gap alone (\(\alpha=1\)), ProD substantially improves over the SFT baseline. However, the best average accuracy is consistently achieved at \(\alpha=0.5\), indicating that balancing both signals provides a more fine-grained estimate of behavioral discrepancies and enables more effective control over optimization strength.

\vspace{-0.05em}
\section{Related Work}
\textbf{Chain-of-Thought Distillation.} Prior works fine-tune student models on complete teacher-generated reasoning chains \citep{fullcotdistill-0, fullcotdistill-1}. To reduce noise from lengthy CoT sequences, several methods extract concise or keypoint-focused rationales via token-weighting modules~\citep{skipcotdistill-0}, reasoning segmentation~\citep{skipcotdistill-1}, or redundant-step pruning~\citep{concisedistill-0, concisedistill-1}. UniCoTT~\citep{unicott} introduces structural-consistency learning for diverse reasoning patterns, while SCOTT~\citep{scott} uses counterfactual training to improve reasoning faithfulness. \citet{reasoningloop} study looping in single-turn CoT distillation and attribute it to hardness-induced risk aversion, highlighting the need for training-time interventions without developing a dedicated mitigation objective. In contrast, we study stagnation in agent distillation, characterize it via the teacher--student stagnation gap, and propose a training-time solution.

\textbf{Agent Distillation.} Distilling agentic capabilities enables SLMs to learn tool-augmented reasoning rather than merely memorize facts, as in conventional CoT distillation. As a result, distilled SLMs exhibit stronger generalization to novel tasks that require previously unseen knowledge. A recent foundational work \citep{agentdistill} establishes an SFT paradigm on teacher-generated interactive trajectories. By masking environmental observations in the training loss, this approach encourages the student to produce coherent thoughts and actions. While effective, it does not adequately address the severe stagnation exhibited by SLMs in interactions, which our method aims to mitigate.

\vspace{-0.2em}
\section{Conclusion}
This work systematically analyzes a key limitation of SFT-based agent distillation: progress stagnation. We find that student agents often become trapped in unproductive, failed strategies, and that this issue worsens with longer interaction horizons and smaller model scales. To address it, we propose Progress-Aware Distillation (ProD), which explicitly detects and adaptively penalizes stagnation by measuring behavioral discrepancies against teacher trajectories. Through iterative optimization, ProD gradually closes the gap toward teacher-level dynamics. Extensive experiments demonstrate that ProD significantly enhances both the task success rates and execution efficiency of SLMs. Future work could develop more fine-grained progress metrics and more precise methods for localizing stagnant actions, enabling targeted corrective signals during agent distillation.

\bibliography{ref}

@article{agent-survey,
  title={Large language model agent: A survey on methodology, applications and challenges},
  author={Luo, Junyu and Zhang, Weizhi and Yuan, Ye and Zhao, Yusheng and Yang, Junwei and Gu, Yiyang and Wu, Bohan and Chen, Binqi and Qiao, Ziyue and Long, Qingqing and others},
  journal={arXiv preprint arXiv:2503.21460},
  year={2025}
}

@article{gpt4,
  title={Gpt-4 technical report},
  author={Achiam, Josh and Adler, Steven and Agarwal, Sandhini and Ahmad, Lama and Akkaya, Ilge and Aleman, Florencia Leoni and Almeida, Diogo and Altenschmidt, Janko and Altman, Sam and Anadkat, Shyamal and others},
  journal={arXiv preprint arXiv:2303.08774},
  year={2023}
}

@misc{gpt4.1,
  title        = {Introducing {GPT-4.1} in the API},
  author       = {{OpenAI}},
  year         = {2025},
  howpublished = {\url{https://openai.com/index/gpt-4-1/}},
  note         = {{Accessed: 2025-12-22}},
}

@article{llama3,
  title={The llama 3 herd of models},
  author={Grattafiori, Aaron and Dubey, Abhimanyu and Jauhri, Abhinav and Pandey, Abhinav and Kadian, Abhishek and Al-Dahle, Ahmad and Letman, Aiesha and Mathur, Akhil and Schelten, Alan and Vaughan, Alex and others},
  journal={arXiv preprint arXiv:2407.21783},
  year={2024}
}

@article{qwen3,
  title={Qwen3 technical report},
  author={Yang, An and Li, Anfeng and Yang, Baosong and Zhang, Beichen and Hui, Binyuan and Zheng, Bo and Yu, Bowen and Gao, Chang and Huang, Chengen and Lv, Chenxu and others},
  journal={arXiv preprint arXiv:2505.09388},
  year={2025}
}

@article{gemma3,
  title={Gemma 3 technical report},
  author={Team, Gemma and Kamath, Aishwarya and Ferret, Johan and Pathak, Shreya and Vieillard, Nino and Merhej, Ramona and Perrin, Sarah and Matejovicova, Tatiana and Ram{\'e}, Alexandre and Rivi{\`e}re, Morgane and others},
  journal={arXiv preprint arXiv:2503.19786},
  year={2025}
}

@article{toolformer,
  title={Toolformer: Language models can teach themselves to use tools},
  author={Schick, Timo and Dwivedi-Yu, Jane and Dess{\`\i}, Roberto and Raileanu, Roberta and Lomeli, Maria and Hambro, Eric and Zettlemoyer, Luke and Cancedda, Nicola and Scialom, Thomas},
  journal={Advances in Neural Information Processing Systems},
  volume={36},
  pages={68539--68551},
  year={2023}
}

@article{webexplorer,
  title={Webexplorer: Explore and evolve for training long-horizon web agents},
  author={Liu, Junteng and Li, Yunji and Zhang, Chi and Li, Jingyang and Chen, Aili and Ji, Ke and Cheng, Weiyu and Wu, Zijia and Du, Chengyu and Xu, Qidi and others},
  journal={arXiv preprint arXiv:2509.06501},
  year={2025}
}

@article{sweagent,
  title={Swe-agent: Agent-computer interfaces enable automated software engineering},
  author={Yang, John and Jimenez, Carlos E and Wettig, Alexander and Lieret, Kilian and Yao, Shunyu and Narasimhan, Karthik and Press, Ofir},
  journal={Advances in Neural Information Processing Systems},
  volume={37},
  pages={50528--50652},
  year={2024}
}

@article{webthinker,
  author       = {Xiaoxi Li and
                  Jiajie Jin and
                  Guanting Dong and
                  Hongjin Qian and
                  Yutao Zhu and
                  Yongkang Wu and
                  Ji{-}Rong Wen and
                  Zhicheng Dou},
  title        = {WebThinker: Empowering Large Reasoning Models with Deep Research Capability},
  journal      = {CoRR},
  volume       = {abs/2504.21776},
  year         = {2025},
  url          = {https://doi.org/10.48550/arXiv.2504.21776},
  doi          = {10.48550/ARXIV.2504.21776},
  eprinttype    = {arXiv},
  eprint       = {2504.21776},
  bibsource    = {dblp computer science bibliography, https://dblp.org}
}

@Misc{deepresearch,
  author       = {{OpenAI}},
  title        = {Introducing Deep Research},
  year         = {2025},
  url          = {https://openai.com/index/introducing-deep-research/},
}

@inproceedings{react,
title={ReAct: Synergizing Reasoning and Acting in Language Models},
author={Shunyu Yao and Jeffrey Zhao and Dian Yu and Nan Du and Izhak Shafran and Karthik R Narasimhan and Yuan Cao},
booktitle={The Eleventh International Conference on Learning Representations },
year={2023},
url={https://openreview.net/forum?id=WE_vluYUL-X}
}

@inproceedings{codeact,
  title={Executable code actions elicit better llm agents},
  author={Wang, Xingyao and Chen, Yangyi and Yuan, Lifan and Zhang, Yizhe and Li, Yunzhu and Peng, Hao and Ji, Heng},
  booktitle={Forty-first International Conference on Machine Learning},
  year={2024}
}

@inproceedings{fullcotdistill-0,
  title={Large language models are reasoning teachers},
  author={Ho, Namgyu and Schmid, Laura and Yun, Se-Young},
  booktitle={Proceedings of the 61st annual meeting of the association for computational linguistics (volume 1: long papers)},
  pages={14852--14882},
  year={2023}
}

@article{fullcotdistill-1,
  title={Distilling reasoning capabilities into smaller language models},
  author={Shridhar, Kumar and Stolfo, Alessandro and Sachan, Mrinmaya},
  journal={Findings of the Association for Computational Linguistics: ACL 2023},
  pages={7059--7073},
  year={2023},
  publisher={Association for Computational Linguistics}
}

@inproceedings{skipcotdistill-0,
  title={Keypoint-based progressive chain-of-thought distillation for LLMs},
  author={Feng, Kaituo and Li, Changsheng and Zhang, Xiaolu and Zhou, Jun and Yuan, Ye and Wang, Guoren},
  booktitle={Proceedings of the 41st International Conference on Machine Learning},
  pages={13241--13255},
  year={2024}
}

@inproceedings{skipcotdistill-1,
  title={Skip-thinking: Chunk-wise chain-of-thought distillation enable smaller language models to reason better and faster},
  author={Chen, Xiaoshu and Zhou, Sihang and Liang, Ke and Sun, Xiaoyu and Liu, Xinwang},
  booktitle={Proceedings of the 2025 Conference on Empirical Methods in Natural Language Processing},
  pages={12153--12168},
  year={2025}
}

@article{concisedistill-0,
  title={Concise Reasoning, Big Gains: Pruning Long Reasoning Trace with Difficulty-Aware Prompting},
  author={Wu, Yifan and Shi, Jingze and Wu, Bingheng and Zhang, Jiayi and Lin, Xiaotian and Tang, Nan and Luo, Yuyu},
  journal={arXiv preprint arXiv:2505.19716},
  year={2025}
}

@article{concisedistill-1,
  title={DRP: Distilled Reasoning Pruning with Skill-aware Step Decomposition for Efficient Large Reasoning Models},
  author={Jiang, Yuxuan and Li, Dawei and Ferraro, Frank},
  journal={arXiv preprint arXiv:2505.13975},
  year={2025}
}

@inproceedings{unicott,
  title={Unicott: A unified framework for structural chain-of-thought distillation},
  author={Zhuang, Xianwei and Zhu, Zhihong and Wang, Zhichang and Cheng, Xuxin and Zou, Yuexian},
  booktitle={The Thirteenth International Conference on Learning Representations},
  year={2025}
}

@inproceedings{scott,
  title={SCOTT: Self-Consistent Chain-of-Thought Distillation},
  author={Wang, Peifeng and Wang, Zhengyang and Li, Zheng and Gao, Yifan and Yin, Bing and Ren, Xiang},
  booktitle={Proceedings of the 61st Annual Meeting of the Association for Computational Linguistics (Volume 1: Long Papers)},
  pages={5546--5558},
  year={2023}
}

@article{agentdistill,
  title={Distilling llm agent into small models with retrieval and code tools},
  author={Kang, Minki and Jeong, Jongwon and Lee, Seanie and Cho, Jaewoong and Hwang, Sung Ju},
  journal={arXiv preprint arXiv:2505.17612},
  year={2025}
}

@Misc{smolagents,
  title =        {`smolagents`: a smol library to build great agentic systems.},
  author =       {Aymeric Roucher and Albert Villanova del Moral and Thomas Wolf and Leandro von Werra and Erik Kaunismäki},
  howpublished = {\url{https://github.com/huggingface/smolagents}},
  year =         {2025}
}

@article{fireact,
  title={Fireact: Toward language agent fine-tuning},
  author={Chen, Baian and Shu, Chang and Shareghi, Ehsan and Collier, Nigel and Narasimhan, Karthik and Yao, Shunyu},
  journal={arXiv preprint arXiv:2310.05915},
  year={2023}
}

@inproceedings{tora,
  title={Tora: A tool-integrated reasoning agent for mathematical problem solving},
  author={Gou, Zhibin and Shao, Zhihong and Gong, Yeyun and Yang, Yujiu and Huang, Minlie and Duan, Nan and Chen, Weizhu and others},
  booktitle={International Conference on Learning Representations},
  volume={2024},
  pages={48362--48395},
  year={2024}
}

@inproceedings{hotpotqa,
  title={HotpotQA: A dataset for diverse, explainable multi-hop question answering},
  author={Yang, Zhilin and Qi, Peng and Zhang, Saizheng and Bengio, Yoshua and Cohen, William and Salakhutdinov, Ruslan and Manning, Christopher D},
  booktitle={Proceedings of the 2018 conference on empirical methods in natural language processing},
  pages={2369--2380},
  year={2018}
}

@inproceedings{MATH,
  title={Measuring Mathematical Problem Solving With the MATH Dataset},
  author={Hendrycks, Dan and Burns, Collin and Kadavath, Saurav and Arora, Akul and Basart, Steven and Tang, Eric and Song, Dawn and Steinhardt, Jacob},
  booktitle={Thirty-fifth Conference on Neural Information Processing Systems Datasets and Benchmarks Track (Round 2)},
 year={2021}
}

@article{lora,
  title={Lora: Low-rank adaptation of large language models.},
  author={Hu, Edward J and Shen, Yelong and Wallis, Phillip and Allen-Zhu, Zeyuan and Li, Yuanzhi and Wang, Shean and Wang, Lu and Chen, Weizhu and others},
  journal={ICLR},
  volume={1},
  number={2},
  pages={3},
  year={2022}
}

@article{simpo,
  title={Simpo: Simple preference optimization with a reference-free reward},
  author={Meng, Yu and Xia, Mengzhou and Chen, Danqi},
  journal={Advances in Neural Information Processing Systems},
  volume={37},
  pages={124198--124235},
  year={2024}
}

@inproceedings{rouge,
    title = "{ROUGE}: A Package for Automatic Evaluation of Summaries",
    author = "Lin, Chin-Yew",
    booktitle = "Text Summarization Branches Out",
    month = jul,
    year = "2004",
    address = "Barcelona, Spain",
    publisher = "Association for Computational Linguistics",
    url = "https://aclanthology.org/W04-1013/",
    pages = "74--81"
}

@article{likelihood-displace,
  title={Unintentional unalignment: Likelihood displacement in direct preference optimization},
  author={Razin, Noam and Malladi, Sadhika and Bhaskar, Adithya and Chen, Danqi and Arora, Sanjeev and Hanin, Boris},
  journal={arXiv preprint arXiv:2410.08847},
  year={2024}
}

@article{search-r1,
  title={Search-r1: Training llms to reason and leverage search engines with reinforcement learning},
  author={Jin, Bowen and Zeng, Hansi and Yue, Zhenrui and Yoon, Jinsung and Arik, Sercan and Wang, Dong and Zamani, Hamed and Han, Jiawei},
  journal={arXiv preprint arXiv:2503.09516},
  year={2025}
}

@inproceedings{omnimath,
  title={Omni-MATH: A Universal Olympiad Level Mathematic Benchmark for Large Language Models},
  author={Gao, Bofei and Song, Feifan and Yang, Zhe and Cai, Zefan and Miao, Yibo and Dong, Qingxiu and Li, Lei and Ma, Chenghao and Chen, Liang and Xu, Runxin and others},
  booktitle={The Thirteenth International Conference on Learning Representations},
  year={2025}
}

@inproceedings{2wiki,
    title = "Constructing A Multi-hop {QA} Dataset for Comprehensive Evaluation of Reasoning Steps",
    author = "Ho, Xanh  and
      Duong Nguyen, Anh-Khoa  and
      Sugawara, Saku  and
      Aizawa, Akiko",
    editor = "Scott, Donia  and
      Bel, Nuria  and
      Zong, Chengqing",
    booktitle = "Proceedings of the 28th International Conference on Computational Linguistics",
    month = dec,
    year = "2020",
    address = "Barcelona, Spain (Online)",
    publisher = "International Committee on Computational Linguistics",
    url = "https://aclanthology.org/2020.coling-main.580/",
    doi = "10.18653/v1/2020.coling-main.580",
    pages = "6609--6625"
}

@article{sentencebert,
  title={Mpnet: Masked and permuted pre-training for language understanding},
  author={Song, Kaitao and Tan, Xu and Qin, Tao and Lu, Jianfeng and Liu, Tie-Yan},
  journal={Advances in neural information processing systems},
  volume={33},
  pages={16857--16867},
  year={2020}
}

@inproceedings{oagent,
    title = "{OA}gents: An Empirical Study of Building Effective Agents",
    author = "Zhu, He  and
      Qin, Tianrui  and
      Zhu, King  and
      Huang, Heyuan  and
      Guan, Yeyi  and
      Xia, Jinxiang  and
      Li, Hanhao  and
      Yao, Yi  and
      Wang, Ningning  and
      Liu, Pai  and
      Peng, Tianhao  and
      Gui, Xin  and
      Xiaowan, Li  and
      Liu, Yuhui  and
      Tang, Xiangru  and
      Yang, Jian  and
      Zhang, Ge  and
      Gao, Xitong  and
      Jiang, Yuchen Eleanor  and
      Zhang, Changwang  and
      Wang, Jun  and
      Liu, Jiaheng  and
      Zhou, Wangchunshu",
    editor = "Christodoulopoulos, Christos  and
      Chakraborty, Tanmoy  and
      Rose, Carolyn  and
      Peng, Violet",
    booktitle = "Findings of the Association for Computational Linguistics: EMNLP 2025",
    month = nov,
    year = "2025",
    address = "Suzhou, China",
    publisher = "Association for Computational Linguistics",
    url = "https://aclanthology.org/2025.findings-emnlp.720/",
    doi = "10.18653/v1/2025.findings-emnlp.720",
    pages = "13354--13369",
    ISBN = "979-8-89176-335-7"
}

@article{keskar2019ctrl,
  title={Ctrl: A conditional transformer language model for controllable generation},
  author={Keskar, Nitish Shirish and McCann, Bryan and Varshney, Lav R and Xiong, Caiming and Socher, Richard},
  journal={arXiv preprint arXiv:1909.05858},
  year={2019}
}

@misc{vllm_repetition_penalty,
  title={{SamplingParams.repetition\_penalty --- vLLM Documentation}},
  author={{vLLM}},
  howpublished={\url{https://docs.vllm.ai/en/latest/api/vllm/\#vllm.SamplingParams.repetition_penalty}},
  year={2026},
  note={Accessed: 2026-05-06}
}

@article{reasoningloop,
  title={Wait, Wait, Wait... Why Do Reasoning Models Loop?},
  author={Pipis, Charilaos and Garg, Shivam and Kontonis, Vasilis and Shrivastava, Vaishnavi and Krishnamurthy, Akshay and Papailiopoulos, Dimitris},
  journal={arXiv preprint arXiv:2512.12895},
  year={2025}
}

@article{dpo,
  title={Direct preference optimization: Your language model is secretly a reward model},
  author={Rafailov, Rafael and Sharma, Archit and Mitchell, Eric and Manning, Christopher D and Ermon, Stefano and Finn, Chelsea},
  journal={Advances in neural information processing systems},
  volume={36},
  pages={53728--53741},
  year={2023}
}

@article{mmlu,
  title={Measuring massive multitask language understanding},
  author={Hendrycks, Dan and Burns, Collin and Basart, Steven and Zou, Andy and Mazeika, Mantas and Song, Dawn and Steinhardt, Jacob},
  journal={arXiv preprint arXiv:2009.03300},
  year={2020}
}

@article{ARC,
  title={Think you have solved question answering? try arc, the ai2 reasoning challenge},
  author={Clark, Peter and Cowhey, Isaac and Etzioni, Oren and Khot, Tushar and Sabharwal, Ashish and Schoenick, Carissa and Tafjord, Oyvind},
  journal={arXiv preprint arXiv:1803.05457},
  year={2018}
}

@inproceedings{hellaswag,
    title = "{H}ella{S}wag: Can a Machine Really Finish Your Sentence?",
    author = "Zellers, Rowan  and
      Holtzman, Ari  and
      Bisk, Yonatan  and
      Farhadi, Ali  and
      Choi, Yejin",
    editor = "Korhonen, Anna  and
      Traum, David  and
      M{\`a}rquez, Llu{\'i}s",
    booktitle = "Proceedings of the 57th Annual Meeting of the {A}ssociation for {C}omputational {L}inguistics",
    month = jul,
    year = "2019",
    address = "Florence, Italy",
    publisher = "Association for Computational Linguistics",
    url = "https://aclanthology.org/P19-1472/",
    doi = "10.18653/v1/P19-1472",
    pages = "4791--4800"
}

@article{winogrande,
  title={Winogrande: An adversarial winograd schema challenge at scale},
  author={Sakaguchi, Keisuke and Bras, Ronan Le and Bhagavatula, Chandra and Choi, Yejin},
  journal={Communications of the ACM},
  volume={64},
  number={9},
  pages={99--106},
  year={2021},
  publisher={ACM New York, NY, USA}
}
\bibliographystyle{abbrvnat}

\newpage
\appendix
\section{LLM-as-Judge Setup for Progress Stagnation} \label{app:judgment}
We use a frontier LLM, \texttt{gpt-5.4-2026-03-05}, to evaluate progress stagnation for both the teacher agent (GPT-4.1) and various student agents. Following the evaluation protocol~\citep{oagent}, which assesses how promising each action in an agent trajectory is for advancing toward the user's task, we ask the judge to assign a stagnation score from \(0.0\) to \(1.0\) starting from the second turn, since the first turn has no prior interaction history for comparison. The full prompt is provided below.

\begin{tcolorbox}[
    enhanced,
    colback=cyan!3,
    colframe=cyan!40!black,
    colbacktitle=cyan!60!black,
    coltitle=white,
    title=Evaluation Prompt for Progress Stagnation,
    fonttitle=\bfseries,
    arc=3pt,
    boxrule=0.8pt,
    top=5pt, bottom=5pt, left=7pt, right=7pt,
    attach boxed title to top left={xshift=2mm,yshift=-2mm},
    boxed title style={
        sharp corners,
        colback=cyan!70!black,
        colframe=cyan!40!black,
        boxrule=0pt,
        top=1pt, bottom=1pt, left=4pt, right=4pt
    }
]
\textbf{System prompt:}\\
You are an expert evaluator of AI agent trajectories. Your task is to assess progress stagnation at each interaction turn.\\

Progress stagnation occurs when an agent repeats previously failed strategies without making meaningful forward progress toward solving the task. A stagnant turn typically:\\
- Issues the same or very similar search queries as prior turns\\
- Executes the same or nearly identical code as before\\
- Repeats reasoning already attempted without introducing new insights or directions\\

For each turn starting from Turn 2, assign a stagnation score from 0.0 to 1.0. Use the full continuous range --- do not cluster around a few values:\\
- 0.0: Completely new direction --- entirely different strategy, query, or reasoning path\\
- 0.2: Mostly new progress --- builds on prior observations with a clearly different approach\\
- 0.4: Slight repetition --- minor overlap with prior turns but introduces meaningful new elements\\
- 0.6: Noticeable stagnation --- similar strategy as before with only superficial variation\\
- 0.8: Strong stagnation --- nearly the same action or reasoning as a prior turn, minimal new content\\
- 1.0: Complete stagnation --- essentially copy-pasting a prior failed action verbatim\\

Intermediate values (e.g.\ 0.1, 0.3, 0.55, 0.75, 0.9) are encouraged for nuanced cases.\\

Return ONLY a valid JSON object with key \texttt{"turn\_scores"} --- a list of floats, one per turn starting from Turn 2.\\

\textbf{User prompt:}\\
Task: \textbf{\{Task description\}}\\

--- Turn 1 ---\\
\textbf{[Thought \& Action]:} \textbf{\{Assistant message at Turn 1\}}\\
\textbf{[Observation]:} \textbf{\{Tool response at Turn 1, truncated to 600 chars\}}\\

--- Turn 2 ---\\
\textbf{[Thought \& Action]:} \textbf{\{Assistant message at Turn 2\}}\\
\textbf{[Observation]:} \textbf{\{Tool response at Turn 2, truncated to 600 chars\}}\\
\textit{...}\\
--- Turn N ---\\
\textbf{[Thought \& Action]:} \textbf{\{Assistant message at Turn N\}}\\
\textbf{[Observation]:} \textbf{\{Tool response at Turn N, truncated to 600 chars\}}\\

Assess stagnation for Turn 2 through Turn N (N-1 score(s) total).\\
Return ONLY JSON: \texttt{\{"turn\_scores": [score\_turn2, score\_turn3, ...]\}}
\end{tcolorbox}

\section{Additional Experimental Details} \label{app:exp_details}
In this section, we present additional implementation details, along with the hyperparameter selection and evaluation prompts used in our experiments. We implement all agents under the CodeAct framework \citep{codeact} using \texttt{smolagents} \citep{smolagents}. The specific version of the teacher agent is \texttt{gpt-4.1-2025-04-14}. All experiments were conducted on a single NVIDIA A100 GPU (80GB) and a single NVIDIA H200 GPU (140GB).

\subsection{Hyperparameters} \label{app:hyper}

\textbf{Hyperparameters in SFT Distillation.} For SFT-based distillation, we first explore a range of learning rates on each task and evaluate model checkpoints from different training epochs on in-distribution evaluation sets to identify the optimal configuration. The selected model is then directly evaluated under out-of-distribution settings. Concretely, we consider learning rates from the set $\{2\times10^{-4},\,2\times10^{-5}\}$, with a batch size of 32, the AdamW optimizer, and a linear learning rate schedule with a 3\% warmup ratio. All models are trained for up to 5 epochs, and checkpoints from each epoch are evaluated. The checkpoint achieving the best accuracy is retained as the final SFT-distilled agent, and all reported results, including stagnation scores, final answer accuracy, and task efficiency, are obtained using these selected models. Across all models and tasks, we find that a learning rate of $2\times10^{-4}$ consistently yields superior performance, which is consistent with prior settings in \citet{agentdistill}. Regarding the number of training epochs, 3 epochs are sufficient for most models to reach optimal performance across tasks. An exception is observed on the MATH benchmark, where Llama3.1-8B and Llama3.2-3B achieve their best results after 2 epochs, and additional training leads to performance degradation.

\textbf{Hyperparameters in Progress-Aware Distillation (ProD).} ProD is applied on top of the SFT-distilled SLMs. We use a batch size of 32, the AdamW optimizer, and a linear learning rate schedule with a 3\% warmup ratio, consistent with the SFT stage. We fix the log-likelihood scaling parameter $\beta$ to 2.5 following \citep{simpo}. For the overall objective \(
\mathcal{L}({\boldsymbol{\theta}}) =   \mathcal{L}_{\text{PRO}}({\boldsymbol{\theta}}) + \lambda \,\mathcal{L}_{\text{SFT}}({\boldsymbol{\theta}})\), we set the weight of the SFT loss term to $\lambda=1$. Empirically, we observe that $\lambda=1$ preserves a high log-probability on teacher trajectories throughout training, and therefore we do not further tune this parameter. For the adaptive margin weighting, we set $\alpha=0.5$ by default and analyze its sensitivity in \cref{sec:ablation}. We select the learning rate from $\{1\times10^{-5},\,5\times10^{-6},\,1\times10^{-6}\}$ and the initial adaptive margin $c_0$ from $\{0.3,\,0.5,\,0.7\}$ based on in-domain evaluation results. \cref{tab:prod_hyper} summarizes the best choices for each model and task. All out-of-domain evaluations are conducted using the selected models without additional training.
\begin{table}[H]
  \caption{Selected learning rates (LR) and initial adaptive margins $c_0$ for ProD.}
  \label{tab:prod_hyper}
  \begin{center}
  \resizebox{0.95\linewidth}{!}{
        \begin{tabular}{c c c c c c c c c c}
\toprule
\textbf{Task} & \textbf{Hyper.}
& \makecell{\textbf{Qwen3}\\\textbf{4B}}
& \makecell{\textbf{Qwen3}\\\textbf{1.7B}}
& \makecell{\textbf{Qwen3}\\\textbf{0.6B}}
& \makecell{\textbf{Llama-3.1}\\\textbf{8B-Inst}}
& \makecell{\textbf{Llama-3.2}\\\textbf{3B-Inst}}
& \makecell{\textbf{Llama-3.2}\\\textbf{1B-Inst}}
& \makecell{\textbf{Gemma-3}\\\textbf{4B-it}}
& \makecell{\textbf{Gemma-3}\\\textbf{1B-it}} \\
\midrule

      \multirow{2}{*}{\textbf{MATH}}
      & LR
      & 5e-6 & 5e-6 & 5e-6 & 5e-6 & 1e-6 & 1e-6 & 5e-6 & 1e-6 \\
      & $c_0$
      & 0.5 & 0.5 & 0.5 & 0.5 & 0.5 & 0.5 & 0.7 & 0.5 \\
      \midrule

      \multirow{2}{*}{\textbf{HotpotQA}}
      & LR
      & 1e-6 & 1e-5 & 1e-6 & 1e-6 & 5e-6 & 5e-6 & 5e-6 & 5e-6 \\
      & $c_0$
      & 0.7 & 0.5 & 0.7 & 0.5 & 0.5 & 0.5 & 0.7 & 0.3 \\
      
      \bottomrule
        \end{tabular}}
  \end{center}
  \vskip -0.1in
\end{table}

\textbf{Hyperparameters in ProD Ablations.} 
For all ProD ablation variants, we follow the same hyperparameter selection procedure as in the full ProD setting. Specifically, the learning rates and fixed margins are selected based on in-domain evaluation performance, and the best-performing configuration for each model and task is reported. The selected hyperparameters for all ablation settings are summarized in \cref{tab:ablation_hyper}. Unless otherwise specified, all other training configurations, including optimizer choice, batch size, learning rate schedule, and loss weights, are kept identical to those used in ProD. For the sensitivity analysis of the adaptive margin weight $\alpha$ reported in \cref{tab:ablation_alpha}, we isolate the effect of $\alpha$ by fixing all other hyperparameters to their default ProD values and varying only $\alpha$. This controlled setup ensures that the observed performance differences can be directly attributed to changes in the adaptive margin weighting.

\begin{table}[H]
  \caption{Selected learning rates (LR) and fixed margins $c_0$ for ProD ablations (w/o SFT+AM and w/o AM) across models and tasks.}
  \label{tab:ablation_hyper}
  \begin{center}
  \resizebox{0.95\linewidth}{!}{
        \begin{tabular}{c c c c c c c c}
      \toprule
\multirow{2}{*}{\raisebox{-0.6ex}{\textbf{Task}}} 
& \multirow{2}{*}{\raisebox{-0.6ex}{\textbf{Hyper.}}}
      & \multicolumn{3}{c}{\textbf{ProD (w/o SFT + AM)}}
      & \multicolumn{3}{c}{\textbf{ProD (w/o AM)}} \\
      \cmidrule(lr){3-5} \cmidrule(lr){6-8}
      & 
      & \textbf{Qwen3-4B}
      & \textbf{Llama-3.1-8-Inst}
      & \textbf{Gemma-3-4B-it}
      & \textbf{Qwen3-4B}
      & \textbf{Llama-3.1-8-Inst}
      & \textbf{Gemma-3-4B-it}\\
      \midrule

      \multirow{2}{*}{\textbf{MATH}}
      & LR
      & 5e-6 & 5e-6 & 1e-6 & 5e-6 & 5e-6 & 5e-6\\
      & $c_0$
      & 0.5 & 0.5 & 0.3 & 0.5 & 0.5 & 0.7\\
      \midrule

      \multirow{2}{*}{\textbf{HotpotQA}}
      & LR
      & 5e-6 & 1e-6 & 5e-6 & 1e-6 & 1e-6 & 5e-6\\
      & $c_0$
      & 0.5 & 0.5 & 0.7 & 0.7 & 0.5 & 0.7\\
      
      \bottomrule
        \end{tabular}}
  \end{center}
  \vskip -0.25in
\end{table}

\subsection{Evaluation Prompts} \label{app:eval_prompt}
For evaluating factual and mathematical reasoning tasks, strict string matching can be brittle and can lead to an inflated false-negative rate. In particular, benign surface-form variations, such as alternative phrasings, equivalent expressions (e.g., decimals vs.\ fractions), or omitted but implied units, are difficult to exhaustively enumerate with hand-crafted normalization rules and therefore leave many corner cases uncovered. Therefore, following prior work \citep{agentdistill, omnimath}, we use an LLM-as-Judge (\texttt{gpt-4.1-mini-2025-04-14}) to assess answer correctness by comparing the model’s final answer against the reference, which is more robust to such formatting variations. We provide the evaluation prompts below.
\begin{tcolorbox}[
    enhanced,
    colback=cyan!3,               
    colframe=cyan!40!black,       
    colbacktitle=cyan!60!black,  
    coltitle=white,               
    title=Evaluation Prompt for Factual Reasoning Task,
    fonttitle=\bfseries,          
    arc=3pt,                      
    boxrule=0.8pt,                
    top=5pt, bottom=5pt, left=7pt, right=7pt,
    attach boxed title to top left={xshift=2mm,yshift=-2mm},
    boxed title style={
        sharp corners,
        colback=cyan!70!black,    
        colframe=cyan!40!black,
        boxrule=0pt,
        top=1pt, bottom=1pt, left=4pt, right=4pt
    }
]
\textbf{User prompt:}\\
Compare the following predicted answer with the gold (correct) answer and determine if they match in meaning given the question. Be reasonable - phrasing differences are okay if the core meaning is the same. You should accept the aliases and the answer conveying the same conclusion. If the predicted answer is overly verbose and fails to capture the key information found in the gold answer (i.e., low recall), consider it a false answer.\\

Question: \textbf{\{Question\}}\\
Predicted answer: \textbf{\{Predicted answer\}}\\
Gold answer: \textbf{\{Ground truth\}}\\

Output your evaluation as a JSON object with two fields:\\
- explanation: Brief explanation of your scoring decision\\
- score: 0 or 1 indicating if answers match
\end{tcolorbox}

\begin{tcolorbox}[
    enhanced,
    colback=cyan!3,               
    colframe=cyan!40!black,       
    colbacktitle=cyan!60!black,  
    coltitle=white,               
    title=Evaluation Prompt for Mathematical Reasoning Task,
    fonttitle=\bfseries,          
    arc=3pt,                      
    boxrule=0.8pt,                
    top=5pt, bottom=5pt, left=7pt, right=7pt,
    attach boxed title to top left={xshift=2mm,yshift=-2mm},
    boxed title style={
        sharp corners,
        colback=cyan!70!black,    
        colframe=cyan!40!black,
        boxrule=0pt,
        top=1pt, bottom=1pt, left=4pt, right=4pt
    }
]
\textbf{User prompt:}\\
Compare the following predicted answer with the correct answer and determine if they are mathematically equivalent..\\

Question: \textbf{\{Question\}}\\
Predicted answer: \textbf{\{Predicted answer\}}\\
Correct answer: \textbf{\{Ground truth\}}\\

Output your evaluation as a JSON object with two fields:\\
- explanation: Brief explanation of your scoring decision\\
- score: 0 or 1 indicating if answers match
\end{tcolorbox}

\section{Evaluation of Progress Stagnation under Full-Parameter Fine-Tuning} \label{app:full_fine_tune}
\begin{figure}[ht]
\begin{center}
\centerline{\includegraphics[width=1.0\linewidth]{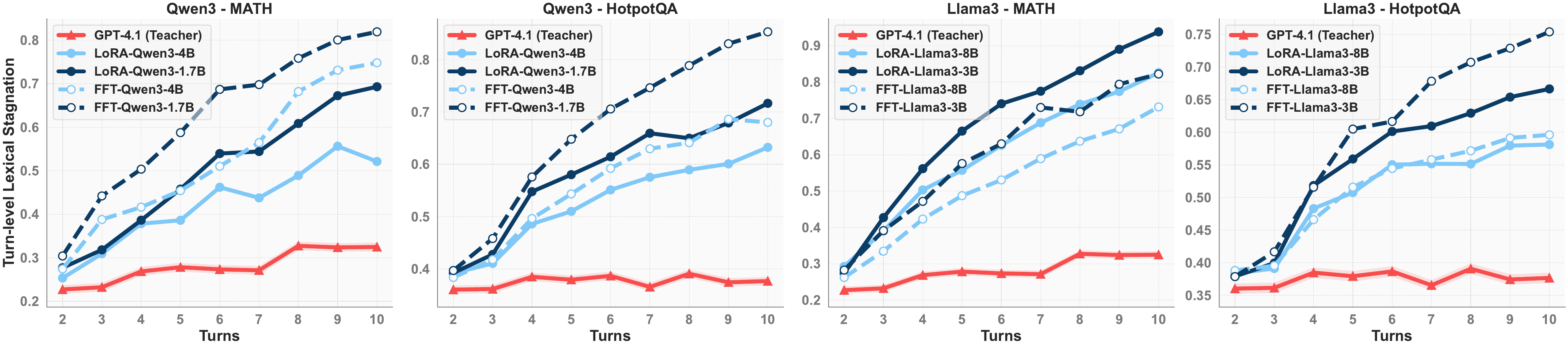}}
\vskip -0.05in
\caption{Comparison of turn-level lexical stagnation scores on MATH and HotpotQA for the teacher agent (GPT-4.1) and SFT-distilled students (Qwen3-4B, Qwen3-1.7B, Llama3.1-8B-Inst, and Llama3.2-3B-Inst) trained with LoRA or full-parameter fine-tuning (FFT).}
\label{fig:full_fine_tuning}
\end{center}
\vskip -0.3in
\end{figure}

In \cref{fig:progress_stagnation} and \cref{fig:mitigate_stagnation}, we have shown that SLMs distilled via LoRA-based SFT exhibit substantially more severe progress stagnation than the teacher agent across a range of tasks and model families. To further investigate the impact of different training paradigms on this phenomenon, we conduct additional experiments using full-parameter fine-tuning (FFT) for SFT-based agent distillation. Following Appendix \ref{app:hyper}, we train FFT for the same number of epochs as the LoRA setting and tune the learning rate over $\{2\times10^{-4},\,2\times10^{-5}\}$, selecting $2\times10^{-5}$ based on in-domain evaluation performance. The resulting stagnation scores are reported in \cref{fig:full_fine_tuning}. We find that FFT-distilled agents still exhibit pronounced progress stagnation, indicating that this failure mode is not alleviated by simply increasing the capacity of parameter updates. Moreover, for the Qwen family, stagnation is sometimes even more severe than under LoRA-based fine-tuning. This suggests that unrestricted parameter updates may amplify imitation bias during the distillation process, thereby exacerbating unproductive agent interaction loops. These findings further underscore the importance of explicitly detecting and penalizing progress stagnation exhibited by the student during optimization, rather than relying solely on imitation-based supervision.

\section{Evaluation of Progress Stagnation on the Distillation Training Data} \label{app:stagnation_training_set}
\begin{figure}[H]
\vskip -0.1in
\begin{center}
\centerline{\includegraphics[width=0.85\linewidth]{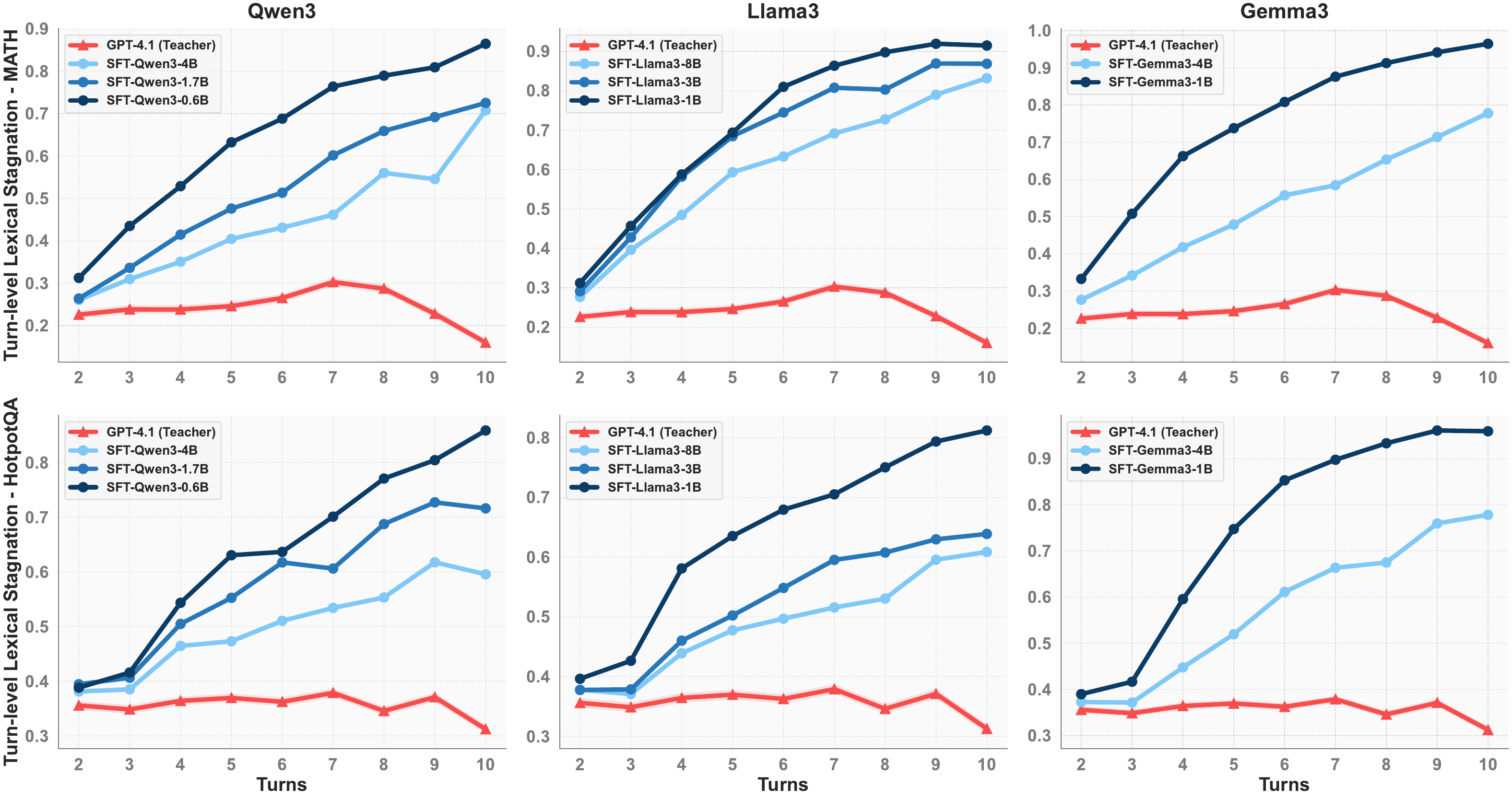}}
\vskip -0.05in
\caption{Comparison of turn-level lexical stagnation scores between the teacher agent (GPT-4.1) and SFT-distilled students on the training dataset.}
\label{fig:stagnation_training_set}
\end{center}
\vskip -0.3in
\end{figure}

In \cref{fig:progress_stagnation} and \cref{fig:mitigate_stagnation}, we demonstrate that SLMs distilled via SFT exhibit pronounced progress stagnation on held-out in-domain evaluation sets. Here, we further show that this stagnation phenomenon persists even on the SFT distillation dataset itself. Specifically, we let the distilled student agent generate its own interaction trajectories in response to user queries drawn from the distillation dataset $\mathcal{D}$, and compute the stagnation score at each interaction turn. \cref{fig:stagnation_training_set} reports the average stagnation scores aggregated over student-generated trajectories from 1,000 sampled user queries, together with the corresponding trajectories produced by the teacher agent. We observe that the student agent exhibits similarly severe stagnation patterns on the training dataset as those observed on held-out evaluation sets. This consistent behavior indicates that imitation-based SFT alone fails to endow student agents with the teacher’s ability to avoid stagnation, even on the data used for distillation. These findings further motivate the design of ProD, which explicitly detects and penalizes stagnation directly on the SFT distillation dataset, enabling the student agent to progressively approximate the teacher’s behavior.

\section{Gradient Analysis of the ProD Objective} \label{app:gradient}
In this section, we analyze the gradient of the progress-aware objective in \cref{eq:prodi} to understand how the adaptive margin modulates the update strength. Our derivation follows the analytical framework of SimPO \citep{simpo}, while introducing a trajectory-dependent margin to capture progress signals:
\begin{equation}
\scalebox{0.92}{$
\nabla_{\boldsymbol{\theta}}\mathcal{L}_{\text{PRO}}(\boldsymbol{\theta})
=
-\beta\,\mathbb{E}_{(\boldsymbol{x}, \tau)\sim\mathcal{D},\,
                 \tilde{\tau}\sim p_S^{\boldsymbol{\theta}_0}(\cdot\mid \boldsymbol{x})}\Big[
w_{\boldsymbol{\theta}}(\tilde{\tau},\tau) \cdot 
\Big(
\frac{1}{|\tau|}\nabla_{\boldsymbol{\theta}}\log p_S^{\boldsymbol{\theta}}(\tau\mid \boldsymbol{x})
-
\frac{1}{|\tilde\tau|}\nabla_{\boldsymbol{\theta}}\log p_S^{\boldsymbol{\theta}}(\tilde\tau\mid \boldsymbol{x})
\Big)
\Big],$}
\end{equation}
where
\begin{equation*}
\scalebox{0.92}{$
w_{\boldsymbol{\theta}}(\tilde{\tau},\tau) = \sigma \Big(   
\frac{\beta}{\lvert \tilde{\tau}\rvert}\,
         \log p_S^{\boldsymbol{\theta}}(\tilde{\tau} \mid \boldsymbol{x})
        -
\frac{\beta}{\lvert \tau\rvert}\,
        \log p_S^{\boldsymbol{\theta}}(\tau \mid \boldsymbol{x})
         + f(\tilde{\tau},\tau) \Big) $}
\end{equation*}
denotes the gradient weight induced by $\mathcal{L}_{\text{PRO}}$. Because the adaptive margin $f(\tilde{\tau},\tau)$ in \cref{eq:margin} accounts for both the stagnation severity of the student trajectory and the correctness of the final outcome, student trajectories that exhibit stronger stagnation and yield incorrect answers are assigned larger margins, which in turn produce larger gradient weights. Conversely, student trajectories that make progress comparable to the teacher and reach correct final answers receive smaller margins and are thus down-weighted during optimization. Consequently, our instance-dependent margin design, driven by the stagnation gap between student and teacher trajectories, focuses updates on unproductive student behaviors while avoiding unnecessary over-correction on trajectories that already exhibit competent progress.

\section{ProD with Alternative Progress Metrics} \label{app:alternative}
In Table~\ref{tab:main_result}, we have shown that ProD trained with the lexical stagnation score substantially outperforms standard SFT distillation across models and tasks. In \cref{sec:progress_stagnation}, we further observed strong agreement among lexical, semantic, and LLM-as-judge stagnation scores. Here, we additionally train ProD using the semantic similarity-based stagnation signal. As shown in Table~\ref{tab:prod_semantic}, ProD-Semantic achieves performance comparable to ProD, confirming that the overall distillation performance is robust to the choice of stagnation metric.
\begin{table}[h]
\centering
\vskip -0.15in
\caption{ProD with alternative progress metrics on HotpotQA.}
\vskip 0.1in
\label{tab:prod_semantic}
\begin{tabular}{lccc}
\toprule
\textbf{Model} & \textbf{SFT} & \textbf{ProD-Semantic} & \textbf{ProD} \\
\midrule
Qwen3-4B    & 66.6 & 69.2 & 70.8 \\
Llama3.2-3B & 61.6 & 66.6 & 67.0 \\
\bottomrule
\end{tabular}
\vskip -0.2in
\end{table}

\section{Additional Results on Gemma 3 Models} \label{app:gemma}
\begin{figure}[H]
\vskip -0.1in
\begin{center}
\centerline{\includegraphics[width=1.0\linewidth]{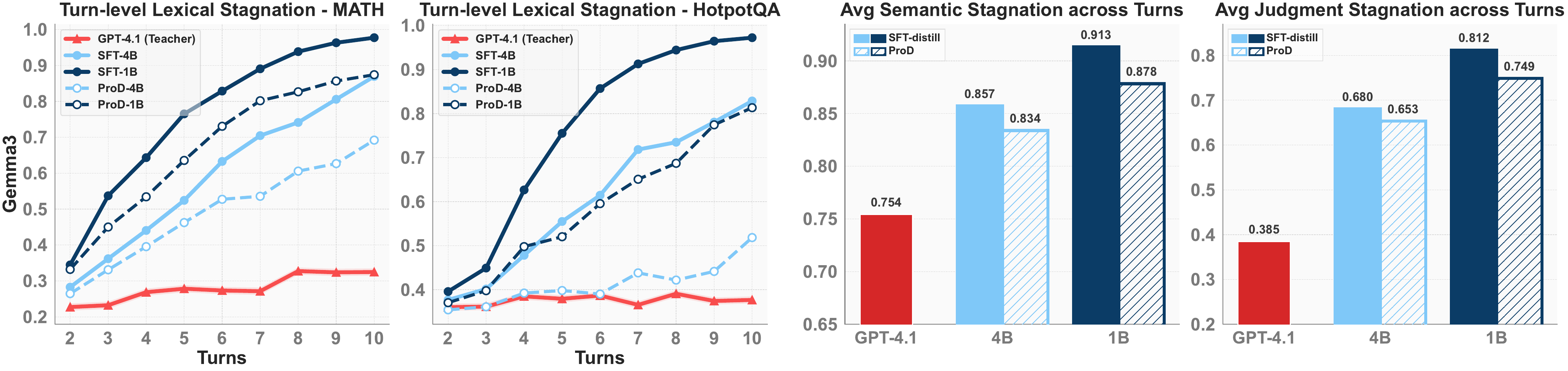}}
\caption{
Alleviating progress stagnation in Gemma3 agents.
Comparison between SFT and ProD:
\textbf{1\&2} show turn-level lexical stagnation on MATH and HotpotQA,
\textbf{3\&4} show average semantic and judgment stagnation across turns and datasets.
}
\label{fig:mitigate_stagnation_gemma3}
\end{center}
\vskip -0.2in
\end{figure}

In \cref{fig:mitigate_stagnation}, we mainly show that ProD substantially alleviates stagnation for Qwen3 and Llama3 agents across lexical, semantic, and LLM-as-judge metrics, enabling SLMs to maintain smoother task progress and better align with teacher behavior.
Here, we provide additional results on Gemma3 models.
As shown in \cref{fig:mitigate_stagnation_gemma3}, ProD similarly reduces the stagnation of Gemma3 student agents, further validating the generality of our method across model families.

\section{Training Overhead} \label{app:train_cost}
ProD is applied on top of SFT, with additional cost mainly arising from student rollout generation, stagnation-score computation, and ProD training, where the main experiments use the lexical variant of the stagnation score. \cref{tab:train_cost} reports the wall-clock time of each component for two representative models, with all experiments conducted on a single H200 GPU.
For each training question, only one student rollout is required, which is generated once as a preprocessing step rather than within the training loop. For 3,000 student trajectories, rollout generation and stagnation-score computation take approximately 1.48~s per trajectory for Qwen3-4B and 0.89~s per trajectory for Qwen3-1.7B, with total times of 73.8 and 44.4 minutes, respectively. These timings reflect sequential rollout generation, which can be parallelized in practice.
Overall, the additional cost remains well controlled.
Moreover, ProD reduces inference token usage by 20.7\%, improving deployment efficiency.

\begin{table}[htb]
\centering
\vskip -0.15in
\caption{Wall-clock time breakdown on HotpotQA (in minutes).}
\label{tab:train_cost}
\vskip 0.1in
\resizebox{0.75\linewidth}{!}{
\begin{tabular}{lcc}
\toprule
\textbf{Component} & \textbf{Qwen3-4B} & \textbf{Qwen3-1.7B} \\
\midrule
SFT training (3 epochs)                                          & 34.7 & 19.7 \\
Rollout generation + stagnation score \textit{(pre-computed)}    & 73.8 & 44.4 \\
ProD training (1 epoch)                                          & 22.2 & 12.1 \\
\bottomrule
\end{tabular}}
\vskip -0.1in
\end{table}

\section{Additional ProD Iterations} \label{app:more_iter}
To examine whether the gains from iterative ProD continue beyond the two iterations studied in \cref{sec:more_iter}, we extend the experiment to four iterations on Llama3.2-3B for MATH. \cref{tab:additional_iter} reports task accuracy and the average turn-level lexical stagnation score. Most of the improvement in both metrics is achieved within the first two iterations. Accuracy peaks at the third iteration, while the stagnation score changes little after the second iteration. Performance remains relatively stable with further iterations, with no observed collapse through the fourth iteration, which still substantially outperforms the SFT baseline. Each additional iteration requires regenerating student rollouts and performing another round of training, so the cumulative cost grows approximately linearly with the number of iterations. Given the limited gains beyond the second iteration in this setting, one or two iterations provide a favorable balance between training cost and performance.

\begin{table}[htb]
\centering
\vskip -0.15in
\caption{Results of additional ProD iterations on Llama3.2-3B for MATH.}
\label{tab:additional_iter}
\vskip 0.1in
\resizebox{0.5\linewidth}{!}{
\begin{tabular}{lcc}
\toprule
\textbf{Method} & \textbf{Accuracy (\%)} $\uparrow$ & \textbf{Lexical Stagnation} $\downarrow$ \\
\midrule
SFT           & 55.5 & 0.678 \\
ProD (Iter-1) & 58.6 & 0.568 \\
ProD (Iter-2) & 61.2 & 0.346 \\
ProD (Iter-3) & \textbf{62.1} & \textbf{0.345} \\
ProD (Iter-4) & 61.0 & 0.348 \\
\bottomrule
\end{tabular}}
\vskip -0.1in
\end{table}

\section{Prompt for Concise-Trajectory SFT} \label{app:concise}
To construct the concise-trajectory SFT baseline, we rewrite long teacher trajectories into shorter demonstrations. Specifically, for teacher trajectories with at least four interaction steps, we prompt the teacher model to solve the same question again using its original successful trajectory as context. The rewriting prompt encourages the model to avoid redundant searches, combine information where possible, and skip unnecessary steps, while still requiring at least one tool-based search. The prompt used for this procedure is shown below.

\begin{tcolorbox}[
    enhanced,
    breakable,
    colback=cyan!3,
    colframe=cyan!40!black,
    colbacktitle=cyan!60!black,
    coltitle=white,
    title=Prompt for Generating Concise Teacher Trajectories,
    fonttitle=\bfseries,
    arc=3pt,
    boxrule=0.8pt,
    top=5pt, bottom=5pt, left=7pt, right=7pt,
    attach boxed title to top left={xshift=2mm,yshift=-2mm},
    boxed title style={
        sharp corners,
        colback=cyan!70!black,
        colframe=cyan!40!black,
        boxrule=0pt,
        top=1pt, bottom=1pt, left=4pt, right=4pt
    }
]
\textbf{User prompt:}\\
A previous attempt correctly solved the following question in \textbf{\{N\}} steps. Previous solution:\\

\textbf{[Step 1]}\\
\textbf{\{Assistant message at Step 1\}}\\

\textbf{[Observation 1]}\\
\textbf{\{Tool response at Step 1\}}\\

\textbf{[Step 2]}\\
\textbf{\{Assistant message at Step 2\}}\\

\textbf{[Observation 2]}\\
\textbf{\{Tool response at Step 2\}}\\
\textit{...}\\
Now solve the same question more concisely. Avoid redundant searches --- combine information where possible and skip unnecessary steps. You must perform at least one search using the tools; do not directly return a final answer without searching.\\

Question: \textbf{\{Question\}}
\end{tcolorbox}

\section{Comparison with Preference-Based Objectives}
In \cref{sec:ablation}, we compare ProD with ProD (w/o SFT + AM), which can be viewed as an on-policy SimPO \citep{simpo} variant using trajectories generated by the student agent. Here, we further compare against on-policy DPO \citep{dpo}. ProD differs from SimPO and DPO by using an adaptive margin explicitly defined by the teacher--student stagnation gap and dynamically adjusted according to trajectory severity. In contrast, SimPO and DPO use fixed-margin or margin-free preference objectives without accounting for trajectory-specific stagnation. As shown in Table~\ref{tab:preference_obj}, both baselines underperform ProD, with DPO falling below SFT on Gemma3-4B. These results demonstrate that the adaptive margin is essential, as non-adaptive preference objectives can either overcorrect or under-optimize trajectories of varying quality.

\begin{table}[h]
\centering
\vskip -0.15in
\caption{On-policy preference optimization baselines, averaged over MATH and HotpotQA.}
\label{tab:preference_obj}
\vskip 0.1in
\resizebox{0.5\linewidth}{!}{
\begin{tabular}{lcccc}
\toprule
\textbf{Model} & \textbf{SFT} & \textbf{DPO} & \textbf{SimPO} & \textbf{ProD} \\
\midrule
Qwen3-4B    & 74.2 & 74.4 & 74.1 & \textbf{78.0} \\
Llama3.1-8B & 61.8 & 62.7 & 64.9 & \textbf{67.2} \\
Gemma3-4B   & 56.4 & 53.8 & 56.7 & \textbf{61.0} \\
\bottomrule
\end{tabular}}
\vskip -0.1in
\end{table}

\section{Comparison with Rejection-Sampling SFT} \label{app:rs_sft}
\begin{figure}[htb]
\centering
\includegraphics[width=1.0\linewidth]{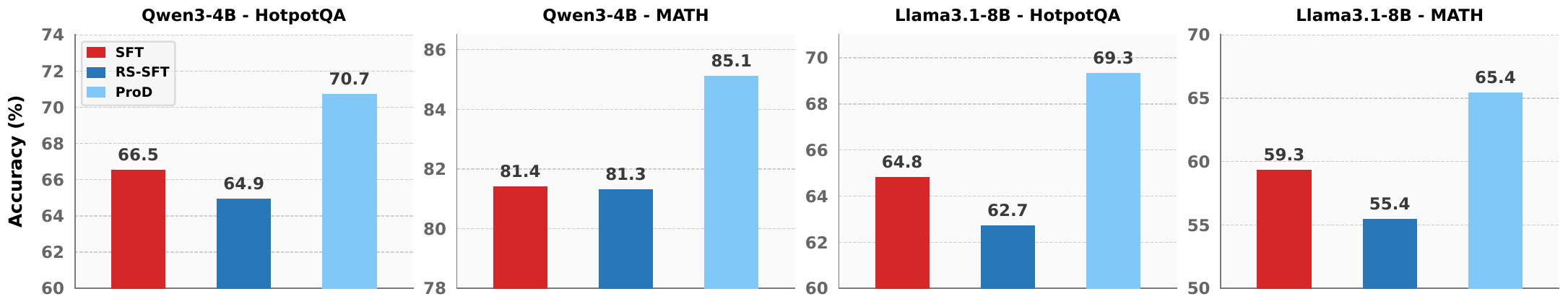}
\vskip -0.05in
\caption{Average task accuracy (\%) for SFT, RS-SFT, and ProD on HotpotQA and MATH.}
\label{fig:rs_sft_accuracy}
\end{figure}

We compare ProD with rejection-sampling SFT (RS-SFT) to examine whether its gains can be obtained by simply retraining on correct student-generated trajectories. Using the same student rollouts as ProD, RS-SFT retains only trajectories with correct final answers and performs additional SFT on them. We tune the learning rate for RS-SFT and report the better-performing setting, keeping all other hyperparameters identical to those used for SFT distillation. \cref{fig:rs_sft_accuracy} reports mean task accuracy over three random seeds for each method. ProD consistently outperforms RS-SFT across both tasks and backbones. Since one iteration of RS-SFT does not improve mean accuracy over the initial SFT model in any of the evaluated settings, we do not perform additional RS-SFT iterations. As shown in \cref{tab:rs_sft_stagnation}, RS-SFT also increases lexical stagnation in all four settings, whereas ProD reduces it. These results support the value of retaining stagnant and failed student trajectories as explicit negative examples during optimization.

\begin{table}[htb]
\centering
\vskip -0.2in
\caption{Average turn-level lexical stagnation scores for SFT, RS-SFT, and ProD.}
\label{tab:rs_sft_stagnation}
\vskip 0.1in
\resizebox{0.55\linewidth}{!}{
\begin{tabular}{llccc}
\toprule
\textbf{Dataset} & \textbf{Model} & \textbf{SFT} & \textbf{RS-SFT} & \textbf{ProD} \\
\midrule
\multirow{2}{*}{HotpotQA} & Qwen3-4B    & 0.544 & 0.574 & \textbf{0.397} \\
         & Llama3.1-8B & 0.525 & 0.550 & \textbf{0.459} \\
\midrule
\multirow{2}{*}{MATH} & Qwen3-4B    & 0.443 & 0.525 & \textbf{0.337} \\
         & Llama3.1-8B & 0.638 & 0.766 & \textbf{0.303} \\
\bottomrule
\end{tabular}}
\vskip -0.1in
\end{table}

\section{Gold-Evidence-Based Progress Analysis} \label{app:gold_evidence}
To complement the lexical, semantic, and LLM-judged stagnation metrics, we further examine whether ProD helps agents acquire new task-relevant evidence during retrieval on HotpotQA. Each question is accompanied by gold supporting facts, which provide human annotations of the evidence required to answer it. We use these annotations to measure evidence-hit rate and additionally examine observation redundancy. Neither metric is used in ProD training.

\textbf{Evidence-hit rate.} We consider a retrieval action progressive if its observation contains a gold supporting-fact passage that has not appeared in any previous observation. Evidence-hit rate is the proportion of retrieval actions that satisfy this condition, measuring how often the agent acquires new evidence relevant to answering the question.

\textbf{Observation redundancy.} As a complementary measure, we compute the proportion of retrieval observations whose passage sets repeat those of earlier observations in the same trajectory. This measures how often the agent revisits previously retrieved content.

As shown in \cref{tab:gold_evidence}, ProD increases evidence-hit rate and reduces observation redundancy across all eight backbones. These results provide additional evidence that ProD promotes useful progress on HotpotQA by retrieving new gold evidence and avoiding repeated retrievals.

\begin{table}[htb]
\centering
\vskip -0.1in
\caption{Evidence-hit rate and observation redundancy (\%) on HotpotQA.}
\label{tab:gold_evidence}
\vskip 0.1in
\resizebox{0.6\linewidth}{!}{
\begin{tabular}{lcccc}
\toprule
\multicolumn{1}{c}{\multirow{2}{*}[-0.6ex]{\textbf{Model}}} & \multicolumn{2}{c}{\textbf{Evidence-hit Rate} $\uparrow$} & \multicolumn{2}{c}{\textbf{Obs. Redundancy} $\downarrow$} \\
\cmidrule(lr){2-3} \cmidrule(lr){4-5}
 & \textbf{SFT} & \textbf{ProD} & \textbf{SFT} & \textbf{ProD} \\
\midrule
Qwen3-4B    & 38.4 & \textbf{44.3} & 14.1 & \textbf{7.0} \\
Qwen3-1.7B  & 31.4 & \textbf{41.1} & 19.4 & \textbf{6.5} \\
Qwen3-0.6B  & 31.3 & \textbf{35.5} & 28.1 & \textbf{14.1} \\
Llama3.1-8B & 36.4 & \textbf{42.5} & 14.6 & \textbf{7.7} \\
Llama3.2-3B & 34.6 & \textbf{43.8} & 16.5 & \textbf{8.0} \\
Llama3.2-1B & 25.5 & \textbf{46.6} & 28.3 & \textbf{5.9} \\
Gemma3-4B   & 35.9 & \textbf{41.6} & 18.4 & \textbf{11.1} \\
Gemma3-1B   & 28.1 & \textbf{47.3} & 45.1 & \textbf{6.7} \\
\bottomrule
\end{tabular}}
\vskip -0.1in
\end{table}

\section{Evaluation of General Capabilities} \label{app:general_capabilities}
To examine the effect of agent distillation on general capabilities, we evaluate Qwen3-4B, Llama3.1-8B, and Gemma3-4B on MMLU \citep{mmlu} (5-shot), ARC-Challenge \citep{ARC} (25-shot), HellaSwag \citep{hellaswag} (10-shot), and WinoGrande \citep{winogrande} (5-shot). We compare the backbone before agent distillation (Base), the SFT-distilled model, and the ProD-distilled model. \cref{tab:general_capabilities} reports accuracy on each benchmark and the average across the four benchmarks. ProD achieves slightly higher average accuracy than both Base and SFT across all three backbones, suggesting that overall performance on these general benchmarks is largely preserved. However, individual benchmarks exhibit small declines. In particular, MMLU accuracy decreases from 72.7 to 71.3 for Qwen3-4B and from 60.3 to 59.5 for Gemma3-4B relative to Base, with both models also falling slightly below SFT. These results indicate that the stable overall performance can coexist with benchmark-specific tradeoffs.

\begin{table}[htb]
\centering
\vskip -0.15in
\caption{General benchmark accuracy (\%) before and after agent distillation.}
\label{tab:general_capabilities}
\vskip 0.1in
\resizebox{0.85\linewidth}{!}{
\begin{tabular}{llccccc}
\toprule
\textbf{Model} & \textbf{Method} & \textbf{MMLU} & \textbf{ARC-C} & \textbf{HellaSwag} & \textbf{WinoGrande} & \textbf{Avg.} \\
\midrule
\multirow{3}{*}{Qwen3-4B}
 & Base & \textbf{72.7} & 62.6 & 62.4 & \textbf{67.2} & 66.2 \\
 & SFT  & 71.8 & 64.0 & 63.0 & 65.8 & 66.2 \\
 & ProD & 71.3 & \textbf{64.8} & \textbf{64.2} & 65.8 & \textbf{66.5} \\
\midrule
\multirow{3}{*}{Llama3.1-8B}
 & Base & \textbf{68.3} & 60.4 & 73.0 & 80.4 & 70.5 \\
 & SFT  & 67.6 & 61.2 & 72.8 & \textbf{82.0} & 70.9 \\
 & ProD & 67.9 & \textbf{63.0} & \textbf{74.0} & 81.6 & \textbf{71.6} \\
\midrule
\multirow{3}{*}{Gemma3-4B}
 & Base & \textbf{60.3} & 59.6 & \textbf{69.6} & 71.6 & 65.3 \\
 & SFT  & 59.9 & 62.2 & 67.8 & 71.2 & 65.3 \\
 & ProD & 59.5 & \textbf{62.8} & 68.6 & \textbf{72.4} & \textbf{65.8} \\
\bottomrule
\end{tabular}}
\vskip -0.1in
\end{table}

\section{Distillation with a Weaker Teacher} \label{app:weaker_teacher}
To examine whether ProD remains effective under a smaller teacher--student capability gap, we use GPT-4o-mini instead of GPT-4.1 as the teacher and Qwen3-1.7B as the student. As shown in \cref{tab:weaker_teacher}, ProD improves task accuracy over SFT from 56.2\% to 60.4\% on HotpotQA and from 63.4\% to 67.5\% on MATH. These results suggest that ProD's gains are preserved with a weaker teacher.

\begin{table}[htb]
\centering
\caption{Task accuracy (\%) with GPT-4o-mini as the teacher and Qwen3-1.7B as the student.}
\label{tab:weaker_teacher}
\vskip 0.1in
\resizebox{0.4\linewidth}{!}{
\begin{tabular}{lccc}
\toprule
\textbf{Dataset} & \textbf{GPT-4o-mini} & \textbf{SFT} & \textbf{ProD} \\
\midrule
HotpotQA & 65.2 & 56.2 & \textbf{60.4} \\
MATH     & 75.2 & 63.4 & \textbf{67.5} \\
\bottomrule
\end{tabular}}
\end{table}

\section{Qualitative Study} \label{app:our_case}
\begin{figure}[ht]
\begin{center}
\centerline{\includegraphics[width=1.0\linewidth]{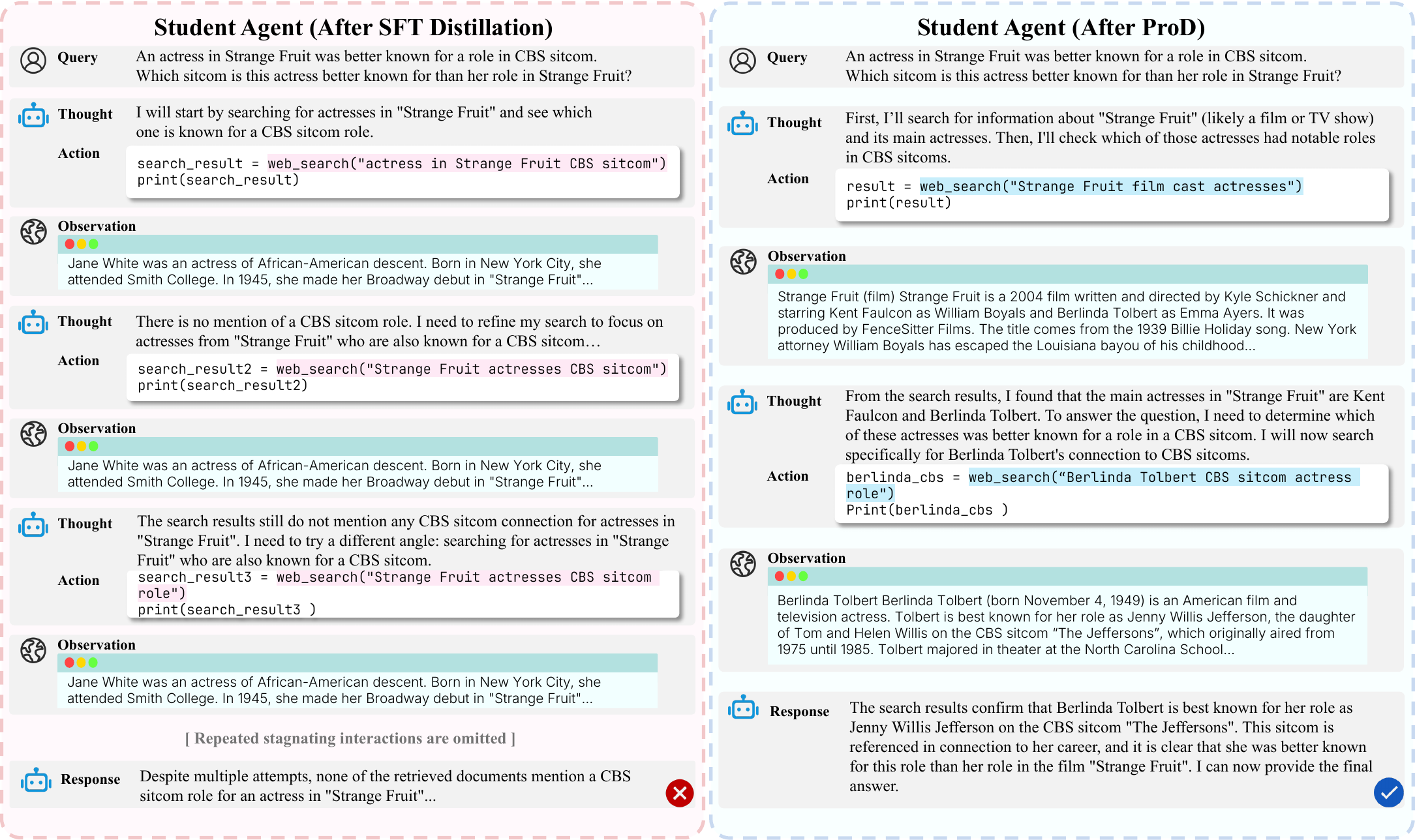}}
\caption{Qualitative comparison between SFT- and ProD-distilled Qwen3-1.7B student agents on a two-hop HotpotQA question. The SFT-distilled agent repeatedly issues similar search queries that attempt to jointly cover both hops, failing to extract useful intermediate information and exhibiting persistent progress stagnation \textbf{(left)}. In contrast, the ProD-distilled agent decomposes the task into a clear two-hop search process, leveraging intermediate observations to guide subsequent actions and achieving steady progress toward the correct answer \textbf{(right)}.}
\label{fig:our_case1}
\end{center}
\vskip -0.2in
\end{figure}

\begin{figure}[ht]
\begin{center}
\centerline{\includegraphics[width=1.0\linewidth]{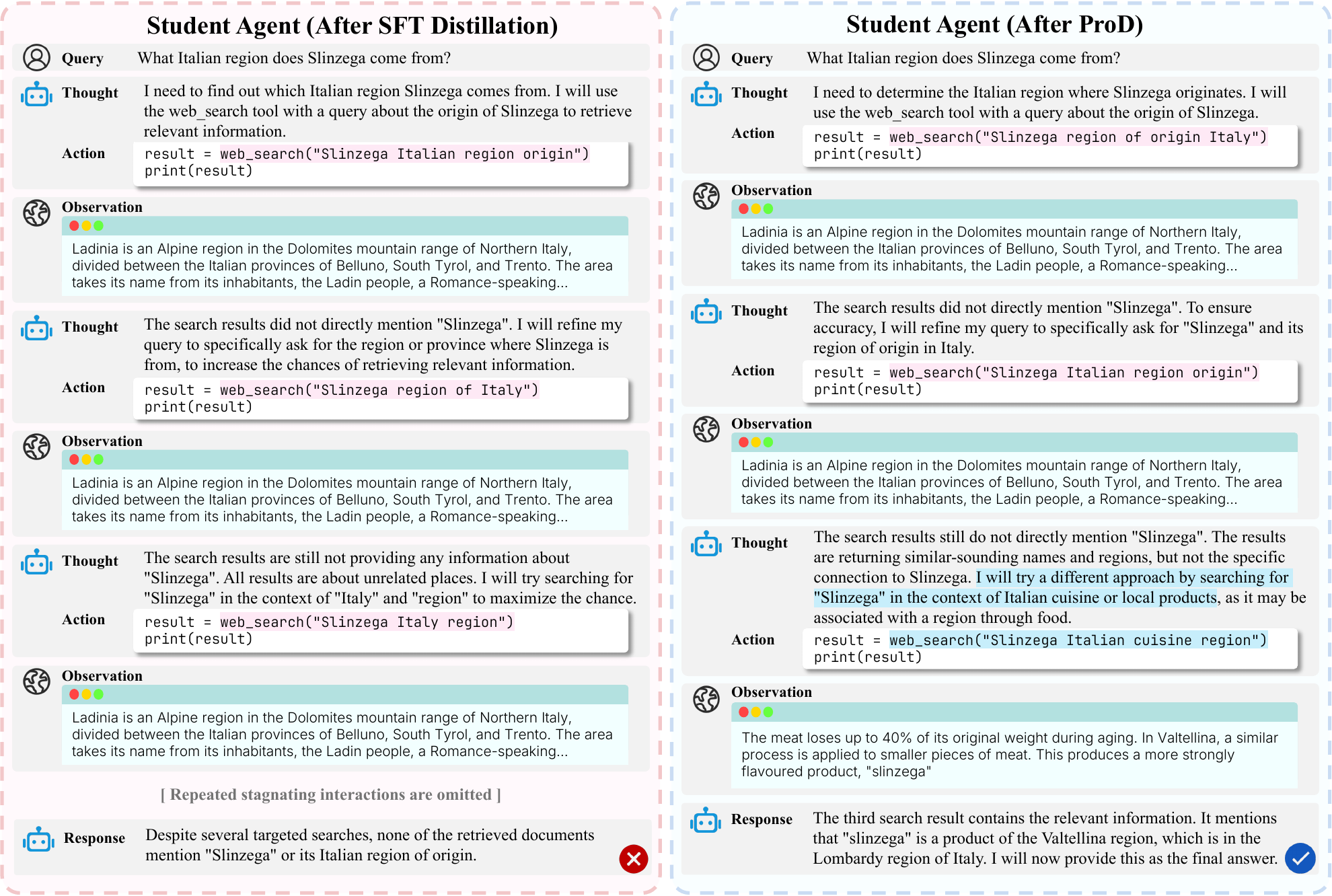}}
\caption{Qualitative comparison between SFT- and ProD-distilled Qwen3-4B student agents on a HotpotQA evaluation example. Although the SFT-distilled agent recognizes the failure of its initial search attempt, it remains trapped in a loop of near-duplicate queries without acquiring new information \textbf{(left)}. By contrast, the ProD-distilled agent adapts its search strategy after early unsuccessful attempts, escapes the stagnation loop, and successfully retrieves the required information to solve the task \textbf{(right)}.}
\label{fig:our_case2}
\end{center}
\vskip -0.2in
\end{figure}

In this section, we present a qualitative analysis of two case studies, each consisting of a paired comparison between interaction trajectories of student agents distilled with SFT and ProD, generated on two sampled questions from the HotpotQA evaluation set. \cref{fig:our_case1} uses the same two-hop query as \cref{fig:case}. We observe that the SFT-distilled agent initially issues a single search query that attempts to cover keywords from both hops simultaneously. When the returned observation does not contain the relevant evidence, the agent continues to persist with highly similar, unsuccessful actions in subsequent turns, failing to acquire new information that would drive progress. In contrast, the ProD-distilled agent follows a solution pattern that closely mirrors the teacher trajectory in \cref{fig:case}: it executes a clear two-hop search procedure and uses intermediate observations to guide subsequent steps, resulting in steady progress toward the answer. \cref{fig:our_case2} exhibits a similar failure mode. Although the SFT-distilled agent recognizes that the first-turn search does not yield useful information, it remains trapped in a loop of near-duplicate search attempts. By comparison, the ProD-distilled agent, after two unsuccessful trials, shifts its search strategy at the third turn in a way that is meaningfully different from the prior failed actions. This change enables the agent to escape the stagnation loop, retrieve the required evidence, and ultimately solve the question. Overall, these examples illustrate that ProD produces student agents with smoother task execution and a stronger ability to break out of unproductive interaction loops compared to SFT distillation.

\clearpage
\section{Discussion} \label{app:discussion}

\textbf{Limitations.}
Our OOD evaluation primarily reflects cross-dataset shifts within the same broad task families. For example, the shift from MATH to Omni-MATH involves moving from high-school competition math to more challenging Olympiad-level problems. Thus, these results should be interpreted as evidence of generalization across datasets within these task families, rather than to entirely new domains. Additionally, stagnation scores should be interpreted as auxiliary diagnostic metrics, while task accuracy remains the primary evaluation metric. Our work focuses on measuring and mitigating stagnation at the trajectory level, where the goal is to quantify how much a student trajectory deviates from the teacher in maintaining progress.
This design is effective for identifying broad teacher--student behavioral gaps, but it does not fully explain the root causes of stagnation. For example, a student may stagnate because it misinterprets an observation, fails to update its plan, over-trusts a failed tool call, or lacks the task knowledge needed to propose a new direction. Our current stagnation scores do not explicitly distinguish these cases. Future work could develop more fine-grained progress metrics that localize stagnation to specific reasoning steps, tool calls, or observation updates, and use this diagnosis to provide more targeted corrective signals during training. 

\textbf{Broader Impact and Safeguard Statement.}
This work improves agent distillation by mitigating progress stagnation in SLM agents, enabling more robust and decisive multi-turn task execution. We expect these advances to support more efficient distillation of reliable autonomous agents from stronger teacher models, especially in settings where large proprietary agents are costly to deploy. At the same time, making agentic capabilities available in smaller and locally deployable models raises important safety considerations, since such agents may interact with external tools, execute code, or retrieve information from the web. To reduce these risks, our current implementation uses the CodeAct framework, which executes code in a sandboxed environment with restricted tool access, providing basic safety guardrails. However, sandboxing alone is not a complete safety solution. When releasing our code and models, we will include usage guidelines and safeguards to discourage misuse.


\clearpage
\section*{NeurIPS Paper Checklist}
\begin{enumerate}

\item {\bf Claims}
    \item[] Question: Do the main claims made in the abstract and introduction accurately reflect the paper's contributions and scope?
    \item[] Answer: \answerYes{} 
    \item[] Justification:  The main claims made in the abstract and introduction accurately reflect the paper’s contributions and scope.

\item {\bf Limitations}
    \item[] Question: Does the paper discuss the limitations of the work performed by the authors?
    \item[] Answer: \answerYes{} 
    \item[] Justification:  The limitations are provided in Appendix \ref{app:discussion}.

\item {\bf Theory assumptions and proofs}
    \item[] Question: For each theoretical result, does the paper provide the full set of assumptions and a complete (and correct) proof?
    \item[] Answer: \answerNA{} 
    \item[] Justification: The paper does not include theoretical results.

    \item {\bf Experimental result reproducibility}
    \item[] Question: Does the paper fully disclose all the information needed to reproduce the main experimental results of the paper to the extent that it affects the main claims and/or conclusions of the paper (regardless of whether the code and data are provided or not)?
    \item[] Answer: \answerYes{} 
    \item[] Justification:  We have provided detailed experimental settings including hyperparameters
needed to reproduce the experimental results in Appendix \ref{app:exp_details}.

\item {\bf Open access to data and code}
    \item[] Question: Does the paper provide open access to the data and code, with sufficient instructions to faithfully reproduce the main experimental results, as described in supplemental material?
    \item[] Answer: \answerYes{} 
    \item[] Justification: Code, data and instructions for reproducing the experiments are publicly available at \url{https://github.com/CaoYuanpu/ProD}.

\item {\bf Experimental setting/details}
    \item[] Question: Does the paper specify all the training and test details (e.g., data splits, hyperparameters, how they were chosen, type of optimizer) necessary to understand the results?
    \item[] Answer: \answerYes{} 
    \item[] Justification: We have provided detailed experimental settings in \cref{sec:exp} and Appendix \ref{app:exp_details}.

\item {\bf Experiment statistical significance}
    \item[] Question: Does the paper report error bars suitably and correctly defined or other appropriate information about the statistical significance of the experiments?
    \item[] Answer: \answerNo{} 
    \item[] Justification: We do not include error bars due to the high computational and evaluation costs involved.

\item {\bf Experiments compute resources}
    \item[] Question: For each experiment, does the paper provide sufficient information on the computer resources (type of compute workers, memory, time of execution) needed to reproduce the experiments?
    \item[] Answer: \answerYes{} 
    \item[] Justification: This is provided in Appendix \ref{app:exp_details}.
    
\item {\bf Code of ethics}
    \item[] Question: Does the research conducted in the paper conform, in every respect, with the NeurIPS Code of Ethics \url{https://neurips.cc/public/EthicsGuidelines}?
    \item[] Answer: \answerYes{} 
    \item[] Justification: The research conducted in the paper conforms with the NeurIPS Code of Ethics.

\item {\bf Broader impacts}
    \item[] Question: Does the paper discuss both potential positive societal impacts and negative societal impacts of the work performed?
    \item[] Answer: \answerYes{} 
    \item[] Justification: We have provided such discussion in Appendix \ref{app:discussion}.
    
\item {\bf Safeguards}
    \item[] Question: Does the paper describe safeguards that have been put in place for responsible release of data or models that have a high risk for misuse (e.g., pre-trained language models, image generators, or scraped datasets)?
    \item[] Answer: \answerYes{} 
    \item[] Justification: We have described safeguards in Appendix \ref{app:discussion}.

\item {\bf Licenses for existing assets}
    \item[] Question: Are the creators or original owners of assets (e.g., code, data, models), used in the paper, properly credited and are the license and terms of use explicitly mentioned and properly respected?
    \item[] Answer: \answerYes{} 
    \item[] Justification:  All the creators and original owners of assets used in the paper are properly credited.

\item {\bf New assets}
    \item[] Question: Are new assets introduced in the paper well documented and is the documentation provided alongside the assets?
    \item[] Answer: \answerYes{}  
    \item[] Justification: The released code is documented at \url{https://github.com/CaoYuanpu/ProD}, including data access and experimental reproduction instructions.

\item {\bf Crowdsourcing and research with human subjects}
    \item[] Question: For crowdsourcing experiments and research with human subjects, does the paper include the full text of instructions given to participants and screenshots, if applicable, as well as details about compensation (if any)? 
    \item[] Answer: \answerNA{} 
    \item[] Justification: The paper does not involve crowdsourcing experiments or research with human subjects.

\item {\bf Institutional review board (IRB) approvals or equivalent for research with human subjects}
    \item[] Question: Does the paper describe potential risks incurred by study participants, whether such risks were disclosed to the subjects, and whether Institutional Review Board (IRB) approvals (or an equivalent approval/review based on the requirements of your country or institution) were obtained?
    \item[] Answer: \answerNA{} 
    \item[] Justification: The paper does not involve crowdsourcing experiments or research with human subjects.

\item {\bf Declaration of LLM usage}
    \item[] Question: Does the paper describe the usage of LLMs if it is an important, original, or non-standard component of the core methods in this research? Note that if the LLM is used only for writing, editing, or formatting purposes and does \emph{not} impact the core methodology, scientific rigor, or originality of the research, declaration is not required.
    \item[] Answer: \answerYes{} 
    \item[] Justification: We use LLMs both as teacher agents for collecting training data and as judges for evaluating task execution outcomes. We provide detailed descriptions of these uses in Appendix~\ref{app:exp_details}.

\end{enumerate}

\end{document}